\documentclass{article}
\usepackage{iclr2027_conference,times}
\usepackage[T1]{fontenc}
\usepackage[utf8]{inputenc}
\usepackage{amsmath,amssymb,amsthm,booktabs,graphicx,microtype}
\usepackage{array,multirow,url,hyperref,placeins}
\newcommand{\method}{\textsc{Spectrum}}
\newcommand{\R}{\mathbb R}

\newcommand{\I}{\mathbf I}
\newcommand{\cA}{\mathcal A}
\newcommand{\cD}{\mathcal D}
\newcommand{\cQ}{\mathcal Q}
\newcommand{\Cbar}{\overline C}
\DeclareMathOperator{\diag}{diag}
\DeclareMathOperator*{\argmin}{arg\,min}
\newtheorem{proposition}{Proposition}

\newcommand{\figfile}[1]{\includegraphics[width=\linewidth]{figures/#1.pdf}}

\hypersetup{colorlinks=true,linkcolor=blue,citecolor=blue,urlcolor=blue,
 pdftitle={SPECTRUM: Proximal Spectral Modulation for Looped Self-Distillation},
 pdfauthor={Yunbo Long, WenJie Chen, Jiaquan Zhang, Guangya Hao, Zihang Zeng, Pengze Li, Xi Chen},
 pdfsubject={Correct-solution diversity in recursive self-improvement through looped self-distillation for code generation},
 pdfkeywords={recursive self-improvement, RSI, looped self-distillation, iterative self-distillation, code generation, solution diversity, proximal spectral modulation}}

\title{SPECTRUM: Proximal Spectral Modulation\\for Looped Self-Distillation}
\author{%
\textbf{Yunbo Long}$^{1}$\quad
\textbf{WenJie Chen}$^{2}$\quad
\textbf{Jiaquan Zhang}$^{2}$\quad
\textbf{Guangya Hao}$^{1}$\\
\textbf{Zihang Zeng}$^{2}$\quad
\textbf{Pengze Li}$^{2}$\quad
\textbf{Xi Chen}$^{2}$\\[0.5ex]
\normalfont $^{1}$University of Cambridge\qquad
\normalfont $^{2}$Fudan University
}
\iclrfinalcopy

\begin{document}
\maketitle
\lhead{Preprint}

\begin{abstract}
A model that learns from its own outputs inherits more than their correctness: it inherits which solutions it produces. We formulate Looped Self-Distillation, a self-evolution framework for code generation in which a model repeatedly generates and learns from its own raw outputs, under a fixed information budget, without ongoing external assessment or test-based selection of the generated samples. We identify a consequential separation: correctness can improve while the breadth of correct implementations contracts. We introduce \method, which re-estimates loss-sensitive key/value geometry from a fixed reference anchor at each round and converts it into full-rank proximal spectral modulation. All generated completions train a single student, whose subsequent inference requires no intervention. After five rounds of experiments on MBPP, \method\ retains $89.9\%$ of the initial model's 64-sample correct AST richness, compared with $66.4\%$ for Vanilla self-distillation and $65.5\%$ for a subspace-projection control. The advantage persists at matched correct-sample counts. Without further training or recalibration, the resulting student also achieves higher matched-correct richness than Vanilla SD on HumanEval+ and APPS Intro, demonstrating transfer of the diversity benefit. These findings establish correct-solution retention as a complementary objective of recursive self-improvement (RSI) and show that generation-time intervention can improve the solution repertoire retained by subsequent students.
\end{abstract}

\section{Introduction}
\label{sec:intro}

Iterative self-distillation has improved rationales and code-data synthesizers \citep{rao2023iterative,zhang2025scoder}. Existing approaches use correctness selection, reward-conditioned revision, evolving teachers, or decoding-based distribution shaping \citep{zelikman2022star,singh2024restem,he2026sdzero,sang2026crisp,zhang2026ssd}. We ask: \emph{what does a model retain when it repeatedly learns from its own outputs without new external assessments of those outputs?}

\begin{figure}[t]
\centering\includegraphics[width=\linewidth]{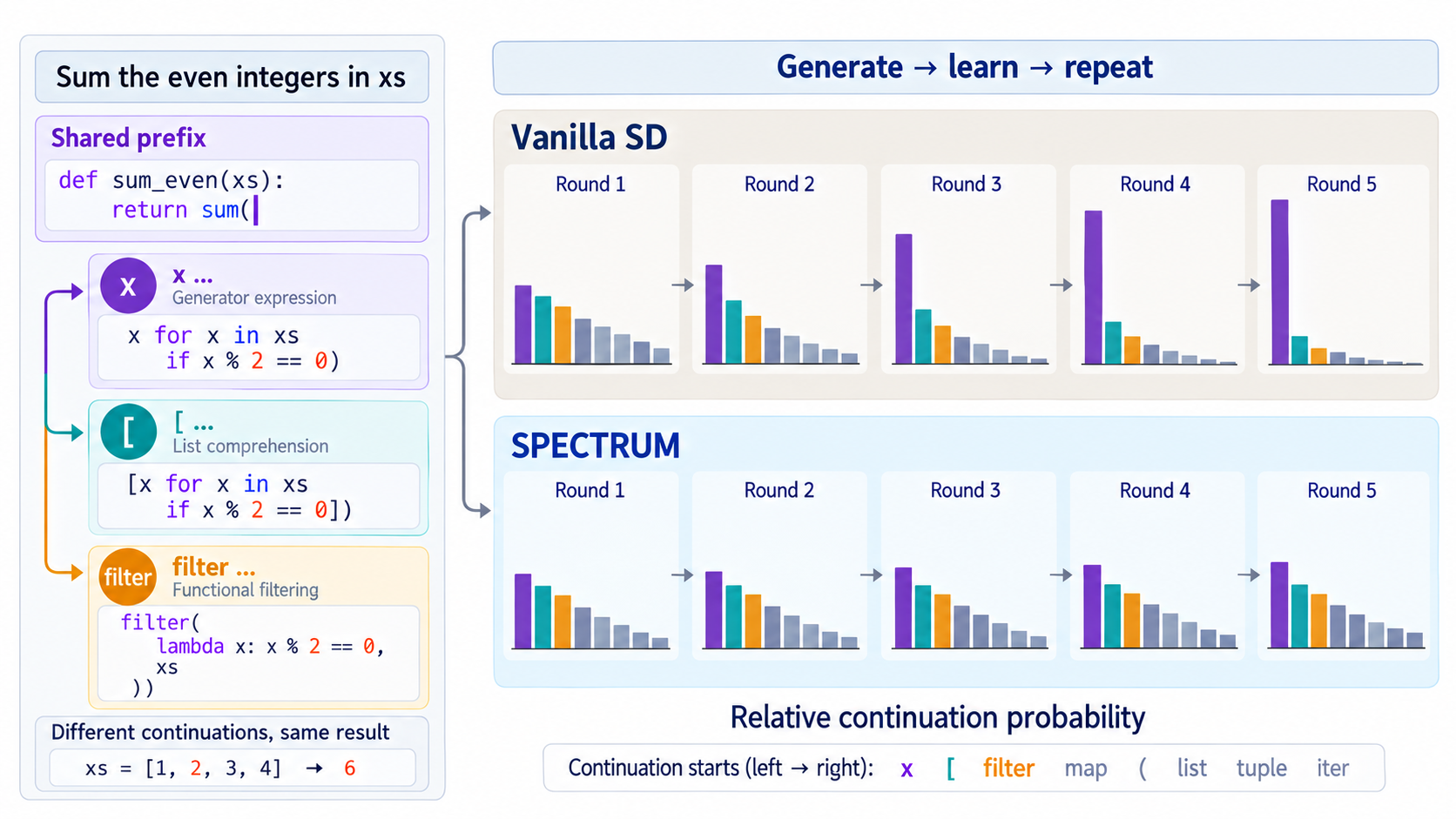}
\caption{\textbf{Each student inherits a distribution of continuations.} A shared code prefix admits different continuation forms. The schematic histograms depict concentration under repeated Vanilla SD and broader retention under \method. Heights are conceptual; the experiments measure complete correct programs, rather than the displayed continuation starts (Figure~\ref{fig:trajectories}).}
\label{fig:motivation}
\end{figure}

Each student supplies its next training distribution. Higher correctness need not make it a better source of diverse correct solutions (Figure~\ref{fig:motivation}). Recursive self-improvement (RSI) therefore calls for measuring whether alternative successful behaviors survive and transfer beyond the training distribution.
We formulate \emph{Looped Self-Distillation under a fixed information budget}. The evolving model generates a corpus, learns from every raw completion, and becomes the next generator. No new annotations, verifier rewards, judge scores, or quality-based sample selection enter training. An optional reference anchor supplies repeated calibration but is never expanded with newly judged samples.
Alternative proofs, plans, and programs make solution diversity relevant across reasoning tasks. For coding agents that use language models to generate candidate programs, repeated self-distillation changes which alternatives those models can supply. We study code generation because executable tests and program structure make solution breadth measurable \citep{chen2021humaneval,liu2023evalplus}, and multiple implementations support program search \citep{li2022alphacode}. Even the entire pass@$k$ curve cannot identify the distribution within correct outputs: redistributing their probability while holding total correctness fixed leaves pass@$k$ unchanged (Proposition~\ref{prop:breadth}).

Our experiments expose a separation between correctness and breadth. Across five rounds, Vanilla SD's pass@1 rises while correct implementation richness contracts (Figure~\ref{fig:trajectories}; Appendix~\ref{app:results}). In the final 64-sample evaluation, pass@1 increases from $37.94\%$ to $41.45\%$, while correct AST richness falls from $12.802$ to $8.506$ classes per task. This connects recursive synthetic-data degradation \citep{shumailov2024collapse,gerstgrasser2024accumulating} and self-distillation diversity loss \citep{nicolicioiu2026diversity} to repeated learning from unselected outputs.

We introduce \method\ to improve retention. Building on reference-gradient calibration \citep{hao2026spd}, it re-estimates the student's loss-sensitive K/V geometry from the same anchor each round. A full-rank proximal gain attenuates less-sensitive directions while retaining a positive local response in every direction. The modulated model generates the corpus; native weights are restored before one LoRA learns from all completions \citep{hu2022lora}.
After five rounds, \method\ retains $89.9\%$ of initial 64-sample richness versus Vanilla SD's $66.4\%$, and improves pass@64 by $2.2$ percentage points over Vanilla SD. Pass@1 is $2.39$ points above initialization and $1.12$ below Vanilla SD. Matched-correct richness gains transfer to HumanEval+ and APPS Intro. Our contributions are:
\begin{enumerate}
\item \textbf{A framework for Looped Self-Distillation.} Repeated learning from raw self-generated corpora under a fixed information budget, without ongoing external assessment or selection of generated samples.
\item \textbf{SPECTRUM: proximal spectral modulation.} Per-round loss-sensitive geometry and a full-rank generation operator transfer a temporary intervention into one native student.
\item \textbf{Solution retention across rounds and benchmarks.} Longitudinal, sampling-budget, generation-design, and transfer analyses establish improved retention within correct implementations.
\end{enumerate}

\section{Preliminaries}
\label{sec:prelim}

\subsection{Self-distillation from model-generated outputs}
SSD learns from self-generated code through cross-entropy training,
using temperature and truncation to shape its synthetic distribution
without correctness-based selection \citep{zhang2026ssd}. We study this
sequence-level learning mechanism when the updated student repeatedly
becomes its own data source.

\subsection{Looped Self-Distillation}
\label{sec:loop}

We formulate \emph{Looped Self-Distillation} as repeated generation
and learning under a fixed information budget. Let $\theta_t$ denote
the current student, $\cQ$ a fixed set of synthesis prompts, and
$\mathcal G$ a generation constructor. One round consists of
\begin{equation}
\begin{aligned}
\widetilde\pi_t
    &= \mathcal G(\theta_t;\cA),\\
\cD_t
    &= \bigl((x,Y_{x,s})\bigr)_{x\in\cQ,\ s=1,\ldots,S},
    \qquad Y_{x,s}\sim\widetilde\pi_t(\cdot\mid x),\\
\theta_{t+1}
    &= \mathcal L(\theta_t,\cD_t),
\end{aligned}
\label{eq:loop}
\end{equation}
where $\cD_t$ is an indexed corpus retaining duplicate records and $\mathcal L$ learns from every generated completion through
cross-entropy training. The optional reference information $\cA$
is fixed before the loop. Newly generated training samples receive
no ongoing external evaluation or supervision: no verifier or judge
determines their selection or training weights.

The loop closes when $\theta_{t+1}$ becomes the next generator, exposing how an update changes the distribution available for future learning. This specifies a learning mechanism within the broader RSI agenda. Our implementation uses one LoRA per round \citep{hu2022lora}; the outer data--parameter loop differs from recurrent-depth transformers that repeat computation within inference \citep{dehghani2019universal}.

\subsection{Correctness and correct-solution breadth}
\label{sec:problem}
For a task $x$, let $v_x(y)\in\{0,1\}$ denote passing its recorded
verifier, and let $\phi(y)$ identify a normalized AST class. Let $p_\theta$ denote raw model likelihood and $\pi_\theta$ the fixed decoding policy induced by the model. Define
$a_\theta(x)=\Pr_{\pi_\theta}[v_x(Y)=1]$ and, when $a_\theta(x)>0$,
\begin{equation}
q_{\theta,j}(x)=\Pr_{\pi_\theta}[\phi(Y)=j\mid v_x(Y)=1],
\qquad
\Pr[v_x(Y)=1,\phi(Y)=j]=a_\theta(x)q_{\theta,j}(x).
\label{eq:factor}
\end{equation}
The first factor measures success frequency; the second describes
the distribution within successful outputs. Suppress $x$ and assume
independent draws from a fixed decoding policy below.

\begin{proposition}[Correctness does not identify correct-solution breadth]
\label{prop:breadth}
For fixed $a>0$, pass@$k=1-(1-a)^k$ is independent of $q$. Expected
correct-class counts under $k$ total draws and $b$ correct draws are
\begin{equation}
C_k(a,q)=\sum_j[1-(1-aq_j)^k],\qquad
D_b(q)=\sum_j[1-(1-q_j)^b].
\label{eq:occupancy}
\end{equation}
Identical pass@$k$ at every $k$ can coexist with different $D_b$ for
any $b\ge2$.
\end{proposition}

We call $C_k$ \emph{correct AST richness@$k$} and $D_b$
\emph{correct-count-matched AST richness@$b$}. Finite-pool rarefaction
estimators \citep{hurlbert1971nonconcept} and the proof appear in
Appendix~\ref{app:metrics}. Conditional comparisons use tasks eligible
for both methods and report their number. Across rounds, relative
richness retention is the ratio of task-mean $C_k$ to its initial-model
value. It measures expected class counts at a fixed budget, rather than
identity overlap with the initial classes.

\section{SPECTRUM: Proximal Spectral Modulation}
\label{sec:method}

\begin{figure}[t]
\centering\includegraphics[width=1\linewidth]{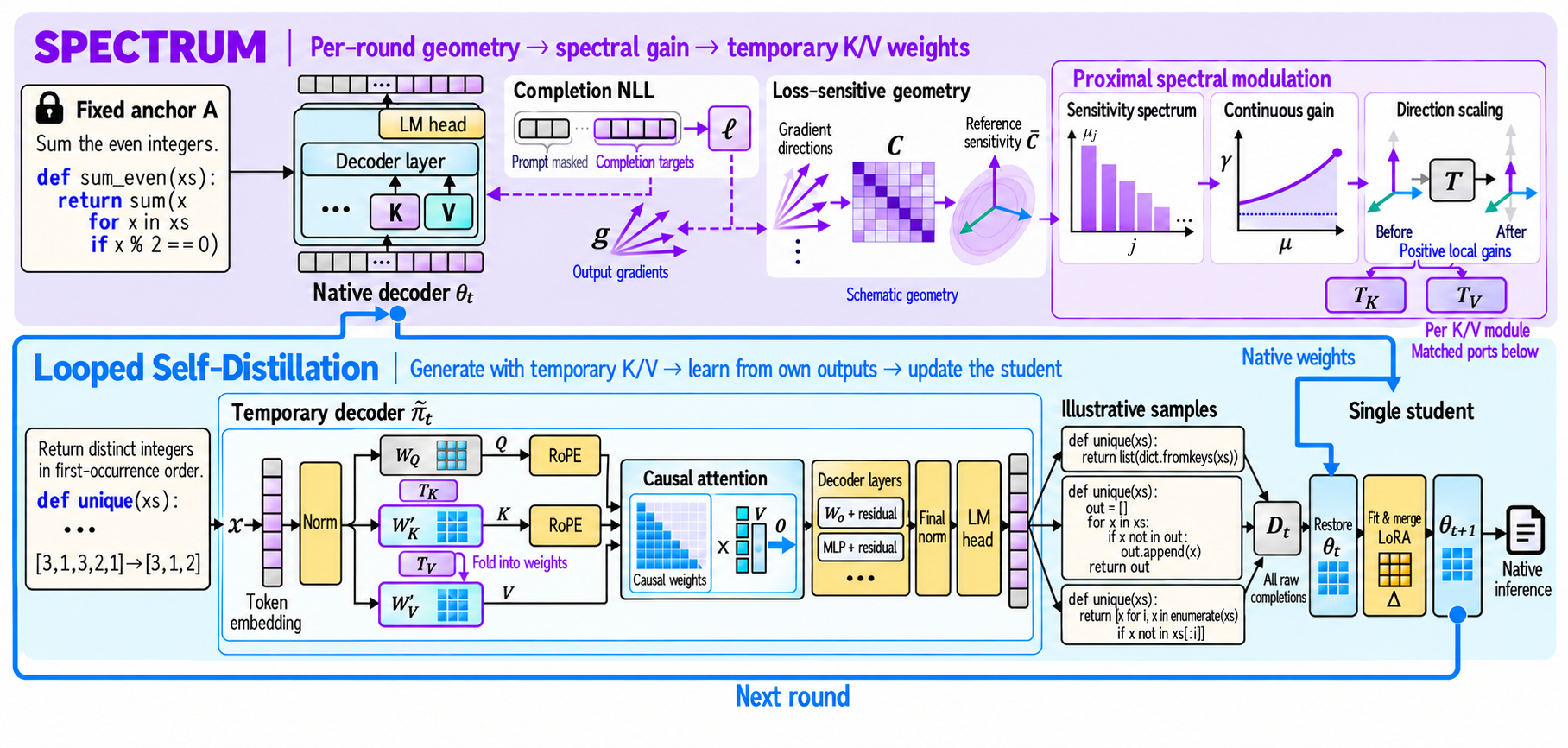}
\caption{\textbf{SPECTRUM within Looped Self-Distillation.} Fixed-anchor completion gradients determine a per-round K/V operator, folded into temporary generation weights. All raw completions train one LoRA student after native weights are restored. The merged student starts the next round and supports native inference. Dashed arrows denote calibration gradients; the bottom loop carries updated parameters. Geometry and code examples are schematic.}
\label{fig:framework}
\end{figure}

\method\ temporarily modulates the current model's K/V representations to generate its next training corpus. Each round recalibrates the geometry from the fixed anchor, generates under a full-rank spectral operator, restores native weights, and trains one student on all raw completions (Figure~\ref{fig:framework}).

\subsection{Loss-sensitive key/value geometry}

Let $\cA=\{(x_i,y_i^\star)\}$ be the reference anchor fixed before the loop. For the current student $\theta_t$, let $\ell_i(\theta_t)$ denote the mean negative log likelihood of reference completion $y_i^\star$. Prompt and padding targets are masked. Following reference-loss-gradient calibration of K/V directions \citep{hao2026spd}, collect the gradient vector $g_{iu}^{(\ell)}\in\R^{d_\ell}$ with respect to the native linear output of each selected K/V module $\ell$ at nonpadding position $u$. Each module is calibrated separately. Its uncentered gradient second moment and normalized geometry are
\begin{equation}
C_{t,\ell}=\frac{1}{M_\ell}\sum_{i,u,\mathrm{nonpad}}
g_{iu}^{(\ell)}g_{iu}^{(\ell)\top},
\qquad
\Cbar_{t,\ell}=\frac{C_{t,\ell}}{\lambda_{\max}(C_{t,\ell})}.
\label{eq:geometry}
\end{equation}
Here $M_\ell$ counts all included positions, including zero gradients. Nonzero calibration signal gives $0\preceq\Cbar_{t,\ell}\preceq\I$. Prompt positions are included because they influence completion loss despite their masked prediction targets.

For a unit direction $v$, $v^\top C_{t,\ell}v$ is the mean squared directional derivative $(g_{iu}^{(\ell)\top}v)^2$. A finite anchor observes only part of the model's behavior, so low measured sensitivity need not imply uselessness for other valid implementations. We use this statistic for continuous attenuation and re-estimate it under each updated student.

\subsection{Full-rank proximal spectral modulation}

Suppressing round and module indices, consider a row activation $h\in\R^{1\times d}$. We choose the quadratic penalty $z(\I-\Cbar)z^\top$ and define
\begin{equation}
\begin{aligned}
z^\star
&=\argmin_{z\in\R^{1\times d}}
\left\{\frac12\|z-h\|_2^2+
\frac{\tau}{2}z(\I-\Cbar)z^\top\right\}
=hT_\tau,\\
T_\tau&=[\I+\tau(\I-\Cbar)]^{-1},\qquad \tau\ge0.
\end{aligned}
\label{eq:prox}
\end{equation}
This quadratic proximal map \citep{parikh2014proximal} limits departure from the native activation while penalizing less-sensitive directions more strongly at fixed norm.

Writing $\Cbar=U\diag(\mu_j)U^\top$ gives the closed-form response
\begin{equation}
T_\tau=U\diag\!\left(\frac{1}{1+\tau(1-\mu_j)}\right)U^\top.
\label{eq:gain}
\end{equation}
The highest-sensitivity directions retain unit gain; all others retain positive gains at finite $\tau$. Strength varies continuously without a prescribed rank.

\begin{samepage}
\begin{proposition}[Local distinguishability and calibration stability]
\label{prop:prox}
Let $0\preceq\Cbar\preceq\I$ and $0\le\tau<\infty$. Equation~\eqref{eq:prox} has a unique minimizer, and the gains of $T_\tau$ lie in $[1/(1+\tau),1]$. For any row-vector difference $\delta$,
\begin{equation}
\frac{\|\delta\|_2}{1+\tau}
\le\|\delta T_\tau\|_2\le\|\delta\|_2,
\qquad
\|T_\tau-\I\|_2\le\frac{\tau}{1+\tau}.
\label{eq:bound}
\end{equation}
For $0\preceq A,B\preceq\I$, the corresponding operators satisfy
$\|T_\tau(A)-T_\tau(B)\|_2\le\tau\|A-B\|_2$.
\end{proposition}
\end{samepage}

The operator also increases or preserves relative alignment with the measured sensitivity geometry. Define $s_{\Cbar}(h)=h\Cbar h^\top/\|h\|_2^2$ for $h\ne0$. Monotonicity of the gains in $\mu_j$ yields
\begin{equation}
s_{\Cbar}(hT_\tau)\ge s_{\Cbar}(h).
\label{eq:relative-sensitivity}
\end{equation}
These local properties establish bounded, sensitivity-aligned modulation and calibration stability (Appendix~\ref{app:proofs}). Retention after attention, decoding, and student learning is measured experimentally.

\subsection{Distilling the modulated generator}

For a native linear module $h=xW^\top+b^\top$, the row operator can be folded into its parameters:
\begin{equation}
W'=T_\tau^\top W,\qquad b'=T_\tau^\top b,
\qquad xW'^\top+b'^\top=hT_\tau.
\label{eq:folding}
\end{equation}
Folding acts at native K/V outputs before rotary transformations and key normalization, requiring no generation-time activation hooks. Keys affect query--key compatibility; values affect the information aggregated by attention.

After generating $\cD_t$, we restore the original weights $\theta_t$ and fit one LoRA adapter \citep{hu2022lora}. Every raw record enters training, including unsuccessful, empty, and duplicate completions. For tokenized record $z=(x,y)$, let $\mathcal T(z)$ contain its nonpadding causal targets, and let $\Delta$ denote the effective LoRA update. The student objective is
\begin{equation}
\mathcal L_t(\Delta)
=-\frac{1}{|\cD_t|}\sum_{z\in\cD_t}
\frac{1}{|\mathcal T(z)|}
\sum_{u\in\mathcal T(z)}
\log p_{\theta_t+\Delta}(z_u\mid z_{<u}).
\label{eq:student}
\end{equation}
SFT includes prompt targets, whereas calibration uses completion targets only. Every record contributes an equally weighted token-mean loss; the operator changes sequence frequencies, not correctness-based training weights.

The merged adapter yields $\theta_{t+1}$. Restoring native weights prevents direct composition of operators across rounds; their influence passes through the corpora. Final inference uses neither operator nor anchor.

\paragraph{Fixed reference information, no ongoing external feedback.}

The reference anchor is supplied before the loop and reused unchanged. Recalibration acquires no new annotations, external-model judgments, or verifier feedback. All generated records retain equal training weight; tests only evaluate the resulting students.

\section{Experiments}
\label{sec:experiments}

\subsection{Protocol and research questions}

\paragraph{Learning and evaluation.}
We study five rounds with Qwen2.5-Coder-1.5B-Instruct \citep{hui2024qwen25coder} on MBPP \citep{austin2021mbpp}. After removing three exact normalized train/test prompt overlaps, the disjoint splits contain 291 synthesis, 50 calibration, 30 validation, and 500 evaluation tasks. Each round generates 16 raw completions per synthesis task and fits one LoRA epoch: 4,656 records per round, 23,280 overall. \method\ uses $\tau=1$, middle and last decoder-layer K/V modules, and LoRA rank 8. The anchor and hyperparameters stay fixed.

Native evaluation uses temperature $0.8$, top-$p$ $0.95$, and no top-$k$ truncation. Every round has 16 samples per task; a separate final pool has 64 for the initial, Vanilla SD, projection, and \method\ checkpoints. SSD is compared under the common 16-sample protocol. We report pass@$k$, $C_k$, and $D_b$, using common eligible tasks for paired conditional contrasts. The longitudinal study uses one training run; intervals use 2,000 task-bootstrap resamples. Appendix~\ref{app:protocol} gives complete settings, and Appendix~\ref{app:matrix} provides independent initial-model context across four models and five benchmarks.

\paragraph{Comparators and questions.}
Vanilla SD uses standard sampling; SSD shapes generation through temperature $1.5$, top-$p$ $0.8$, and top-$k$ 20 \citep{zhang2026ssd}. A top-half rank projector is a generation-design ablation adapted from \citet{hao2026spd}. Projection and \method\ share the 50 references and calibration procedure; Vanilla SD and SSD use no anchor. Synthesis tasks, record counts, and optimization schedules are shared. We ask what survives successive updates (\textbf{RQ1}), whether gains persist within correct outputs (\textbf{RQ2}), how generation design changes the result (\textbf{RQ3}), and whether retention transfers (\textbf{RQ4}). A separate one-round study supplies complementary evidence under a different training budget.

\subsection{RQ1: What do successive students retain?}
\label{sec:rq1}

\begin{figure}[t]
\centering\includegraphics[width=\linewidth]{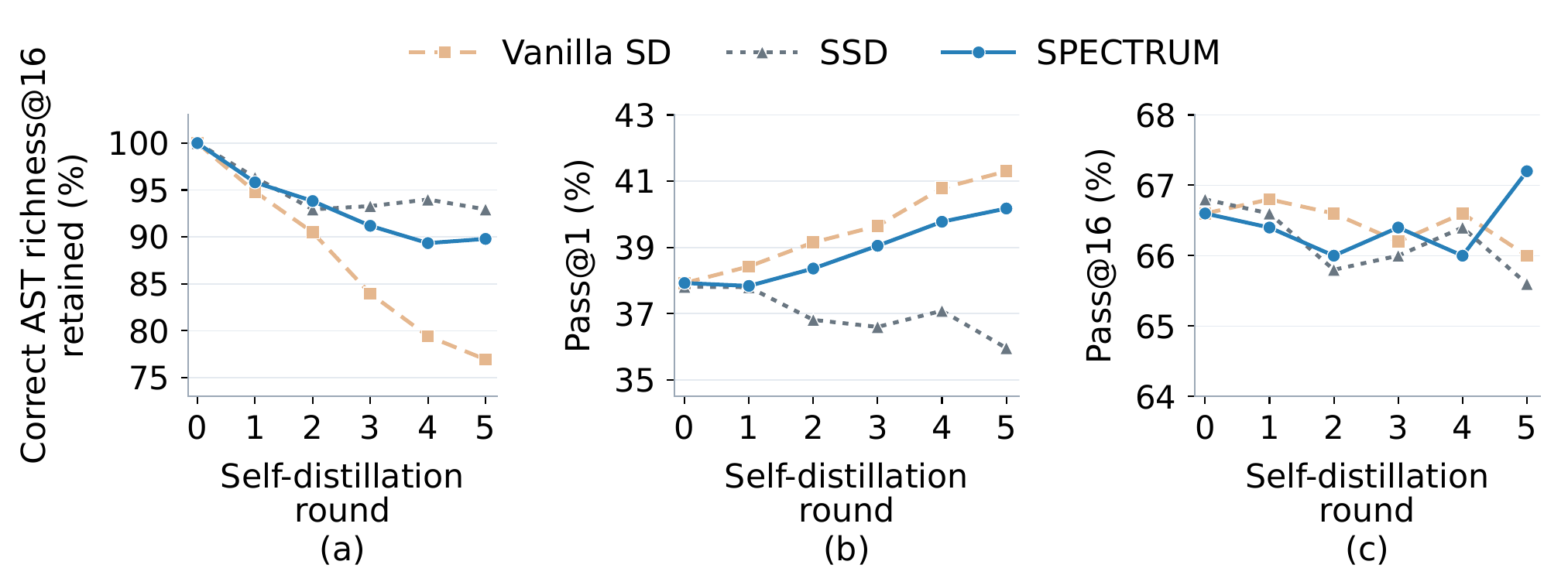}
\caption{\textbf{Generation design separates correctness from solution retention.} Native students on 500 MBPP tasks, 16 samples per task per round. (a) $C_{16}$ relative to each recorded initialization; (b) pass@1; (c) pass@16. All panels show point estimates from one training trajectory per design. SSD's initial model is sampled independently. Additional round-level results appear in Appendix~\ref{app:results}.}
\label{fig:trajectories}
\end{figure}

Vanilla SD's pass@1 rises from $37.92\%$ to $41.30\%$ while $C_{16}$ falls from $4.016$ to $3.090$. \method\ instead retains $89.8\%$ of initial $C_{16}$, versus Vanilla SD's $76.9\%$, with greater richness at every learned round (Figure~\ref{fig:trajectories}a). Its correctness also improves: pass@1 ends at $40.18\%$, and pass@16 at $67.2\%$ versus Vanilla SD's $66.0\%$. The joint trajectory identifies a student that improves success while retaining more of its correct repertoire.

\begin{table}[t]
\centering\small
\caption{\textbf{Final native students after five MBPP rounds.} Independent 64-sample evaluation on 500 tasks. Bold marks the better trained-student estimate. $D_4$ is a marginal mean over 322/321/323 eligible tasks, respectively; paired contrasts follow separately.}
\label{tab:main}
\begin{tabular}{@{}lrrr@{}}
\toprule
Metric $\uparrow$ & Initial model & Vanilla SD & \method\\
\midrule
pass@1 (\%) &37.94&\textbf{41.45}&40.33\\
pass@8 (\%) &61.10&61.86&\textbf{61.89}\\
pass@16 (\%) &65.69&65.64&\textbf{66.23}\\
pass@64 (\%) &72.00&70.40&\textbf{72.60}\\
Correct AST richness@64 ($C_{64}$) &12.802&8.506&\textbf{11.510}\\
Relative richness retention (\%) &100.0&66.4&\textbf{89.9}\\
Correct-count-matched richness@4 ($D_4$) &3.358&2.856&\textbf{3.143}\\
\bottomrule
\end{tabular}
\end{table}

The separate 64-sample pool confirms the retention advantage (Table~\ref{tab:main}): $11.510$ correct AST classes versus $8.506$, or $35.3\%$ more, with paired difference $3.004$ $[2.568,3.494]$. Retention reaches $89.9\%$ versus $66.4\%$. \method\ solves 363 tasks versus 352, a pass@64 gain of $2.2$ percentage points $[0.4,4.2]$. Pass@1 is $2.39$ points above initialization and $1.12$ below Vanilla SD $[-1.61,-0.63]$. Thus richer native students solve more tasks at high budget, with a smaller single-draw success rate than Vanilla SD.

\subsection{RQ2: Does the advantage persist within correct outputs?}
\label{sec:rq2}

\begin{figure}[t]
\centering\includegraphics[width=\linewidth]{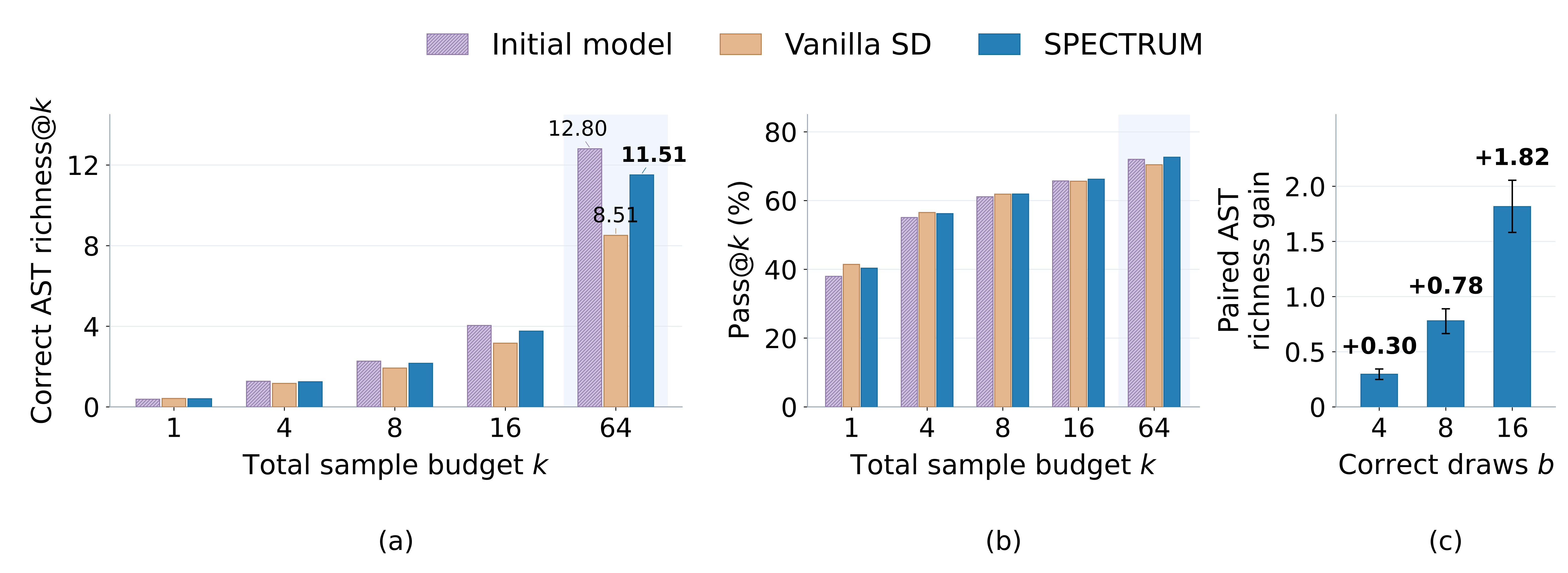}
\caption{(a,b) Correct AST richness and pass@$k$ from the same final 64-sample pools on 500 MBPP tasks. Budget groups are categorical. (c) Paired \method-minus-Vanilla SD $D_b$ gains; whiskers are 95\% task-bootstrap intervals on 316, 296, and 254 common eligible tasks for $b=4,8,16$.}
\label{fig:budget}
\end{figure}

Matching successful draws isolates breadth within correct outputs. At four correct draws, \method\ exceeds Vanilla SD by $0.295$ expected AST classes $[0.249,0.343]$ on 316 common eligible tasks. At eight and sixteen correct draws, the paired gains are $0.780$ and $1.816$ on 296 and 254 tasks (Figure~\ref{fig:budget}c). The advantage therefore persists after equalizing correct-sample counts; each budget uses its own common eligible population.

The total-budget curves provide the complementary view. At $k=1$, $C_1$ equals correctness and favors Vanilla SD. At every displayed larger budget, \method\ produces more distinct correct structures, with its largest pass@$k$ advantage at $k=64$ (Figure~\ref{fig:budget}a,b). These endpoints distinguish breadth bought by a generation budget from variety conditional on success.

\subsection{RQ3: How does generation design affect retention and correctness?}
\label{sec:rq3}

\begin{table}[t]
\centering\small
\setlength{\tabcolsep}{4pt}
\caption{\textbf{Generation designs and realized training tokens.} Round-five native students, 500 MBPP tasks, 16 evaluation draws. Success and richness are higher-is-better; $D_4$ uses each method's eligible tasks (Appendix~\ref{app:results}). SFT tokens sum five rounds, in millions. All designs use 23,280 training records; these are not matched-compute measurements.}
\label{tab:design}
\begin{tabular}{@{}lrrrrrr@{}}
\toprule
Design & pass@1 (\%) & pass@16 (\%) & $C_8$ & $C_{16}$ & $D_4$ & \shortstack{SFT tokens\\(M)}\\
\midrule
Vanilla SD &41.30&66.00&1.902&3.090&2.747&6.373\\
SSD &35.96&65.60&2.091&3.710&3.268&7.326\\
Projection control &41.35&65.60&1.876&3.066&2.690&6.240\\
\method &40.18&67.20&2.091&3.606&2.997&6.856\\
\bottomrule
\end{tabular}
\end{table}

\paragraph{Improving correctness relative to decoding-based SSD.}
\method\ achieves the highest pass@16 among the evaluated designs. Across the five-round trajectories, its pass@1 improves by $2.25$ points from its recorded initialization, whereas SSD declines by $1.85$ points (Figure~\ref{fig:trajectories}b). At round five, \method\ exceeds SSD by $4.21$ points in pass@1 and $1.60$ in pass@16. Their $C_8$ point estimates are nearly identical ($2.091$), while SSD has higher $C_{16}$ and marginal $D_4$ (Table~\ref{tab:design}). The designs occupy distinct correctness--richness operating points.

The accuracy gain uses $8.1\%$ fewer generated completion tokens and $6.4\%$ fewer SFT tokens than SSD. Relative to Vanilla SD, generated tokens are $9.9\%$ higher. These are realized text budgets, not wall-clock or total-compute comparisons: \method\ also performs calibration (Appendix~\ref{app:resources}).

\begin{samepage}
\paragraph{Continuous modulation versus projection.}
With the same reference anchor, calibration, module locations, and learning schedule, continuous modulation retains $89.9\%$ of initial $C_{64}$ versus projection's $65.5\%$. It gains $3.122$ classes and $2.4$ pass@64 points, with pass@1 lower by $1.43$ points. On 251 common eligible tasks in the 16-sample pool, paired $D_4$ gains are $0.312$ $[0.253,0.377]$ for exact programs and $0.293$ $[0.233,0.353]$ for AST classes; the coarser control-flow gain is $0.042$ $[-0.011,0.095]$. The evidence is strongest for implementation breadth. Attenuation strengths differ, so this contrast evaluates the complete operators.

\end{samepage}

\subsection{RQ4: Does retained diversity transfer across benchmarks?}
\label{sec:rq4}

We freeze the MBPP-trained final students and evaluate 164 HumanEval+ tasks and a fixed 200-task APPS Intro subset with 16 samples per task, without further training or recalibration. HumanEval+ uses EvalPlus 0.3.1 \citep{liu2023evalplus}; APPS uses the recorded Python function-interface adaptation \citep{hendrycks2021apps} (Appendix~\ref{app:protocol}).

\begin{table}[t]
\centering\small
\setlength{\tabcolsep}{5pt}
\caption{\textbf{Matched-correct richness transfers to both benchmarks.} \method\ minus Vanilla SD after MBPP learning. $D_4$ uses common eligible tasks; accuracy differences use all tasks. Intervals are 95\% paired task-bootstrap intervals. Full absolute results appear in Table~\ref{tab:transfer-full}.}
\label{tab:transfer}
\begin{tabular}{@{}lrrrrr@{}}
\toprule
Benchmark & $\Delta D_4\uparrow$ & 95\% interval & Common tasks & \shortstack{$\Delta$pass@1\\(pp)} & \shortstack{$\Delta$pass@16\\(pp)}\\
\midrule
HumanEval+ & $+0.080$ & $[0.021,0.142]$ &111&$-1.52$&$0.00$\\
APPS Intro & $+0.104$ & $[0.007,0.205]$ &42&$-1.22$&$-1.50$\\
\bottomrule
\end{tabular}
\end{table}

The paired richness advantage is positive on both benchmarks (Table~\ref{tab:transfer}). On HumanEval+, \method\ also matches Vanilla SD's pass@16 ($86.59\%$) and raises the $C_{16}$ point estimate from $5.848$ to $6.000$. APPS $C_{16}$ falls from $1.930$ to $1.845$, alongside its lower accuracy. Thus the transferable benefit lies in matched-correct implementation breadth, with benchmark-dependent success rates. Appendix~\ref{app:cases} shows recorded equal-correct-count examples.

\subsection{Complementary evidence from a one-round study}
A one-round study evaluates 64 MBPP tasks with 64 samples each. It uses one generate--learn cycle, 128 synthesis tasks with one completion each, and five SFT epochs, separately from the five-round protocol. \method\ improves SSD's pass@32, pass@64, and matched-correct richness (Table~\ref{tab:pilot}); the recorded paired $D_4$ gain is $0.0312$ $[0.0095,0.0550]$, with pass@1 lower by $0.244$ points. This supplies complementary evidence under another training budget. Appendix~\ref{app:pilot} reports all four designs and the code-extraction and verification protocol.

\begin{table}[t]
\centering\small
\caption{\textbf{One-round comparison with SSD.} 64 MBPP tasks, 64 samples per task; a separate training protocol. Values are the archived rounded summaries. Bold marks the larger estimate. Full comparisons and reported intervals are in Appendix~\ref{app:pilot}.}
\label{tab:pilot}
\begin{tabular}{@{}lrrrrrr@{}}
\toprule
Design & pass@1 (\%) & pass@8 (\%) & pass@32 (\%) & pass@64 (\%) & $D_4$ & \shortstack{AST\\Simpson}\\
\midrule
SSD &\textbf{36.89}&\textbf{56.79}&61.66&62.50&3.391&0.863\\
\method &36.65&56.18&\textbf{63.11}&\textbf{65.63}&\textbf{3.422}&\textbf{0.872}\\
\bottomrule
\end{tabular}
\end{table}

\section{Related work}
\label{sec:related}

\paragraph{Sequence-level self-distillation.}
SSD fits self-generated, unverified code after shaping its distribution through temperature and truncation \citep{zhang2026ssd}. This is our closest learning paradigm. \method\ instead shapes generation with fixed reference-loss geometry and measures what native students retain across complete generation--learning cycles. Vanilla SD and the adapted SSD recipe use no reference completions.

\paragraph{Recursive self-improvement and evolving teachers.}
STaR and ReST-EM learn from solutions selected through correctness feedback \citep{zelikman2022star,singh2024restem}. Iterative self-distillation has explicit precedents: \citet{rao2023iterative} use critics and filtering for contexts and rationales, while SCoder improves a code-data synthesizer through checkpoint sampling, scoring, and selection \citep{zhang2025scoder}. SD-Zero distills reward-conditioned self-revision with teacher synchronization \citep{he2026sdzero}; CRISP refreshes an instruction-conditioned teacher for token-level KL training without correctness filtering \citep{sang2026crisp}. Our setting fixes synthesis prompts and optional reference information, retains every raw completion, and examines successive students as future data sources.

\paragraph{Diversity under recursive learning.}
\citet{nicolicioiu2026diversity} show functional and semantic diversity loss with sampled correct demonstrations supplied to a self-teacher. Our loop omits that demonstration-conditioned feedback. UA-RL rewards rare strategies \citep{hu2026uarlacl}; \method\ modulates generation without scoring or labeling generated strategies. Recursive model-collapse studies examine synthetic-data degradation, and data accumulation can mitigate it \citep{shumailov2024collapse,gerstgrasser2024accumulating}. We hold the historical-data policy fixed and jointly measure pass@$k$, $C_k$, and $D_b$ to separate correctness from breadth within correct outputs.

\section{Limitations}
\label{sec:limitations}
Our evaluation has three limitations. First, AST richness captures implementation structure rather than semantic algorithm identity. Structurally different programs may share an underlying algorithm, so semantic and behavioral comparisons would provide a more complete account of solution diversity. Second, finite tests and sample budgets can miss invalid behaviors and rare implementations, motivating broader behavioral checks and adaptive sampling. Third, diversity requires task-specific interpretation: Python AST classes do not directly correspond to distinct proofs, plans, or explanations across languages. Extending this framework requires representations that capture meaningful alternatives within each domain.

\section{Conclusion and future work}
\label{sec:conclusion}
We introduce Looped Self-Distillation as a framework for repeated self-improvement under a fixed information budget, without ongoing external assessment of generated samples. Each student supplies its next training distribution, and correctness can improve while correct-solution breadth contracts. \method\ addresses this retention problem through evolving K/V geometry and full-rank proximal modulation, transferring a temporary generation intervention into a native student through unfiltered data. After five MBPP rounds, it retains $89.9\%$ of initial 64-sample richness versus Vanilla SD's $66.4\%$. Matched-correct richness gains also appear on HumanEval+ and APPS Intro.

Retention and transfer matter for long-horizon RSI. Future work will investigate whether lifelong agents can consolidate experience into reusable concepts and strategies, then recombine them to solve unfamiliar problems without additional task-specific training.
\clearpage
% \section*{Reproducibility statement}
% The appendix specifies the metric estimators, proofs, information flow, losses, data splits, prompts, decoding, and resource accounting. Supplementary tables report complete round-level results, matched-correct comparisons, transfer effects, and resource use. Additional analyses use saved task-level results; training and evaluation protocols are documented separately from these analyses. Code: \url{https://github.com/Yunbo-max/SPECTRUM}.

% \section*{AI use statement}
% AI assistants supported literature retrieval, experiment design and software, mathematical derivation and checking, result analysis, manuscript drafting, and figure preparation. The studied models generated the synthetic training corpora as specified in the protocol. Reported values come from recorded evaluations and identified analyses of saved outputs.

% \subsection*{Ethics statement}
% This study uses existing public coding benchmarks and involves no newly recruited human participants. Passing finite tests does not establish that generated code is safe or suitable for deployment; deployment requires independent validation. 

\bibliographystyle{iclr2027_conference}
\bibliography{references}

@misc{hao2026spd,
  title = {Self-Policy Distillation via Capability-Selective Subspace Projection},
  author = {Hao, Guangya and Shang, Yitong and Long, Yunbo and Zhao, Zhuokai and Liang, Hanxue},
  year = {2026},
  eprint = {2605.22675},
  archivePrefix = {arXiv},
  primaryClass = {cs.CL},
  note = {arXiv preprint},
  url = {https://arxiv.org/abs/2605.22675},
  doi = {10.48550/arXiv.2605.22675}
}

@misc{zhang2026ssd,
  title = {Embarrassingly Simple Self-Distillation Improves Code Generation},
  author = {Zhang, Ruixiang and Bai, Richard He and Zheng, Huangjie and Jaitly, Navdeep and Collobert, Ronan and Zhang, Yizhe},
  year = {2026},
  eprint = {2604.01193},
  archivePrefix = {arXiv},
  primaryClass = {cs.CL},
  note = {arXiv preprint},
  url = {https://arxiv.org/abs/2604.01193},
  doi = {10.48550/arXiv.2604.01193}
}

@article{shumailov2024collapse,
  title = {{AI} Models Collapse When Trained on Recursively Generated Data},
  author = {Shumailov, Ilia and Shumaylov, Zakhar and Zhao, Yiren and Papernot, Nicolas and Anderson, Ross and Gal, Yarin},
  journal = {Nature},
  year = {2024},
  volume = {631},
  pages = {755--759},
  doi = {10.1038/s41586-024-07566-y},
  url = {https://www.nature.com/articles/s41586-024-07566-y}
}

@misc{gerstgrasser2024accumulating,
  title = {Is Model Collapse Inevitable? Breaking the Curse of Recursion by Accumulating Real and Synthetic Data},
  author = {Gerstgrasser, Matthias and Schaeffer, Rylan and Dey, Apratim and Rafailov, Rafael and Sleight, Henry and Hughes, John and Korbak, Tomasz and Agrawal, Rajashree and Pai, Dhruv and Gromov, Andrey and Roberts, Daniel A. and Yang, Diyi and Donoho, David L. and Koyejo, Sanmi},
  year = {2024},
  eprint = {2404.01413},
  archivePrefix = {arXiv},
  primaryClass = {cs.LG},
  note = {arXiv preprint},
  url = {https://arxiv.org/abs/2404.01413},
  doi = {10.48550/arXiv.2404.01413}
}

@inproceedings{hu2022lora,
  title = {{LoRA}: Low-Rank Adaptation of Large Language Models},
  author = {Hu, Edward J. and Shen, Yelong and Wallis, Phillip and Allen-Zhu, Zeyuan and Li, Yuanzhi and Wang, Shean and Wang, Lu and Chen, Weizhu},
  booktitle = {International Conference on Learning Representations},
  year = {2022},
  url = {https://openreview.net/forum?id=nZeVKeeFYf9},
  eprint = {2106.09685},
  archivePrefix = {arXiv}
}

@inproceedings{liu2023evalplus,
  title = {Is Your Code Generated by {ChatGPT} Really Correct? Rigorous Evaluation of Large Language Models for Code Generation},
  author = {Liu, Jiawei and Xia, Chunqiu Steven and Wang, Yuyao and Zhang, Lingming},
  booktitle = {Advances in Neural Information Processing Systems},
  volume = {36},
  year = {2023},
  url = {https://proceedings.neurips.cc/paper_files/paper/2023/hash/43e9d647ccd3e4b7b5baab53f0368686-Abstract-Conference.html},
  eprint = {2305.01210},
  archivePrefix = {arXiv}
}

@misc{austin2021mbpp,
  title = {Program Synthesis with Large Language Models},
  author = {Austin, Jacob and Odena, Augustus and Nye, Maxwell and Bosma, Maarten and Michalewski, Henryk and Dohan, David and Jiang, Ellen and Cai, Carrie and Terry, Michael and Le, Quoc and Sutton, Charles},
  year = {2021},
  eprint = {2108.07732},
  archivePrefix = {arXiv},
  primaryClass = {cs.PL},
  note = {arXiv preprint},
  url = {https://arxiv.org/abs/2108.07732},
  doi = {10.48550/arXiv.2108.07732}
}

@misc{chen2021humaneval,
  title = {Evaluating Large Language Models Trained on Code},
  author = {Chen, Mark and Tworek, Jerry and Jun, Heewoo and Yuan, Qiming and {Ponde de Oliveira Pinto}, Henrique and Kaplan, Jared and Edwards, Harri and Burda, Yuri and Joseph, Nicholas and Brockman, Greg and Ray, Alex and Puri, Raul and Krueger, Gretchen and Petrov, Michael and Khlaaf, Heidy and Sastry, Girish and Mishkin, Pamela and Chan, Brooke and Gray, Scott and Ryder, Nick and Pavlov, Mikhail and Power, Alethea and Kaiser, Lukasz and Bavarian, Mohammad and Winter, Clemens and Tillet, Philippe and Such, Felipe Petroski and Cummings, Dave and Plappert, Matthias and Chantzis, Fotios and Barnes, Elizabeth and Herbert-Voss, Ariel and Guss, William Hebgen and Nichol, Alex and Paino, Alex and Tezak, Nikolas and Tang, Jie and Babuschkin, Igor and Balaji, Suchir and Jain, Shantanu and Saunders, William and Hesse, Christopher and Carr, Andrew N. and Leike, Jan and Achiam, Josh and Misra, Vedant and Morikawa, Evan and Radford, Alec and Knight, Matthew and Brundage, Miles and Murati, Mira and Mayer, Katie and Welinder, Peter and McGrew, Bob and Amodei, Dario and McCandlish, Sam and Sutskever, Ilya and Zaremba, Wojciech},
  year = {2021},
  eprint = {2107.03374},
  archivePrefix = {arXiv},
  primaryClass = {cs.LG},
  note = {arXiv preprint},
  url = {https://arxiv.org/abs/2107.03374},
  doi = {10.48550/arXiv.2107.03374}
}

@misc{hui2024qwen25coder,
  title = {{Qwen2.5-Coder} Technical Report},
  author = {Hui, Binyuan and Yang, Jian and Cui, Zeyu and Yang, Jiaxi and Liu, Dayiheng and Zhang, Lei and Liu, Tianyu and Zhang, Jiajun and Yu, Bowen and Lu, Keming and Dang, Kai and Fan, Yang and Zhang, Yichang and Yang, An and Men, Rui and Huang, Fei and Zheng, Bo and Miao, Yibo and Quan, Shanghaoran and Feng, Yunlong and Ren, Xingzhang and Ren, Xuancheng and Zhou, Jingren and Lin, Junyang},
  year = {2024},
  eprint = {2409.12186},
  archivePrefix = {arXiv},
  primaryClass = {cs.CL},
  note = {arXiv preprint},
  url = {https://arxiv.org/abs/2409.12186},
  doi = {10.48550/arXiv.2409.12186}
}

@article{parikh2014proximal,
  title = {Proximal Algorithms},
  author = {Parikh, Neal and Boyd, Stephen},
  journal = {Foundations and Trends in Optimization},
  volume = {1},
  number = {3},
  year = {2014},
  doi = {10.1561/2400000003},
  url = {https://web.stanford.edu/~boyd/papers/prox_algs.html}
}

@inproceedings{hu2026uarlacl,
  title = {Rewarding the Rare: Uniqueness-Aware {RL} for Creative Problem Solving in {LLM}s},
  author = {Hu, Zhiyuan and Wang, Yucheng and He, Yufei and Wu, Jiaying and Zhao, Yilun and Ng, See-Kiong and Breazeal, Cynthia and Luu, Anh Tuan and Park, Hae Won and Hooi, Bryan},
  booktitle = {Findings of the Association for Computational Linguistics: ACL 2026},
  year = {2026},
  month = jul,
  pages = {39765--39790},
  address = {San Diego, California, United States},
  publisher = {Association for Computational Linguistics},
  doi = {10.18653/v1/2026.findings-acl.1982},
  url = {https://aclanthology.org/2026.findings-acl.1982/}
}

@misc{zelikman2022star,
  title = {{STaR}: Bootstrapping Reasoning With Reasoning},
  author = {Zelikman, Eric and Wu, Yuhuai and Mu, Jesse and Goodman, Noah D.},
  year = {2022},
  eprint = {2203.14465},
  archivePrefix = {arXiv},
  primaryClass = {cs.LG},
  url = {https://arxiv.org/abs/2203.14465}
}

@article{singh2024restem,
  title = {Beyond Human Data: Scaling Self-Training for Problem-Solving with Language Models},
  author = {Singh, Avi and Co-Reyes, John D. and Agarwal, Rishabh and Anand, Ankesh and Patil, Piyush and Garcia, Xavier and Liu, Peter J. and Harrison, James and Lee, Jaehoon and Xu, Kelvin and Parisi, Aaron and Kumar, Abhishek and Alemi, Alex and Rizkowsky, Alex and Nova, Azade and Adlam, Ben and Bohnet, Bernd and Elsayed, Gamaleldin and Sedghi, Hanie and Mordatch, Igor and Simpson, Isabelle and Gur, Izzeddin and Snoek, Jasper and Pennington, Jeffrey and Hron, Jiri and Kenealy, Kathleen and Swersky, Kevin and Mahajan, Kshiteej and Culp, Laura and Xiao, Lechao and Bileschi, Maxwell L. and Constant, Noah and Novak, Roman and Liu, Rosanne and Warkentin, Tris and Qian, Yundi and Bansal, Yamini and Dyer, Ethan and Neyshabur, Behnam and Sohl-Dickstein, Jascha and Fiedel, Noah},
  journal = {Transactions on Machine Learning Research},
  year = {2024},
  eprint = {2312.06585},
  archivePrefix = {arXiv},
  url = {https://arxiv.org/abs/2312.06585}
}

@misc{he2026sdzero,
  title = {Self-Distillation Zero: Self-Revision Turns Binary Rewards into Dense Supervision},
  author = {He, Yinghui and Kaur, Simran and Bhaskar, Adithya and Yang, Yongjin and Liu, Jiarui and Ri, Narutatsu and Fowl, Liam and Panigrahi, Abhishek and Chen, Danqi and Arora, Sanjeev},
  year = {2026},
  eprint = {2604.12002},
  archivePrefix = {arXiv},
  primaryClass = {cs.CL},
  url = {https://arxiv.org/abs/2604.12002}
}

@article{hurlbert1971nonconcept,
  title = {The Nonconcept of Species Diversity: A Critique and Alternative Parameters},
  author = {Hurlbert, Stuart H.},
  journal = {Ecology},
  year = {1971},
  volume = {52},
  number = {4},
  pages = {577--586},
  doi = {10.2307/1934145},
  url = {https://esajournals.onlinelibrary.wiley.com/doi/10.2307/1934145}
}

@inproceedings{dehghani2019universal,
  title = {Universal Transformers},
  author = {Dehghani, Mostafa and Gouws, Stephan and Vinyals, Oriol and Uszkoreit, Jakob and Kaiser, Lukasz},
  booktitle = {International Conference on Learning Representations},
  year = {2019},
  eprint = {1807.03819},
  archivePrefix = {arXiv},
  url = {https://arxiv.org/abs/1807.03819}
}

@misc{hendrycks2021apps,
  title = {Measuring Coding Challenge Competence With {APPS}},
  author = {Hendrycks, Dan and Basart, Steven and Kadavath, Saurav and Mazeika, Mantas and Arora, Akul and Guo, Ethan and Burns, Collin and Puranik, Samir and He, Horace and Song, Dawn and Steinhardt, Jacob},
  year = {2021},
  eprint = {2105.09938},
  archivePrefix = {arXiv},
  primaryClass = {cs.SE},
  url = {https://arxiv.org/abs/2105.09938}
}

@misc{li2022alphacode,
  title = {Competition-Level Code Generation with {AlphaCode}},
  author = {Li, Yujia and Choi, David and Chung, Junyoung and Kushman, Nate and Schrittwieser, Julian and Leblond, R{\'e}mi and Eccles, Tom and Keeling, James and Gimeno, Felix and Dal Lago, Agustin and Hubert, Thomas and Choy, Peter and de Masson d'Autume, Cyprien and Babuschkin, Igor and Chen, Xinyun and Huang, Po-Sen and Welbl, Johannes and Gowal, Sven and Cherepanov, Alexey and Molloy, James and Mankowitz, Daniel J. and Sutherland Robson, Esme and Kohli, Pushmeet and de Freitas, Nando and Kavukcuoglu, Koray and Vinyals, Oriol},
  year = {2022},
  eprint = {2203.07814},
  archivePrefix = {arXiv},
  primaryClass = {cs.PL},
  doi = {10.1126/science.abq1158},
  url = {https://arxiv.org/abs/2203.07814}
}

@misc{jain2024livecodebench,
  title = {{LiveCodeBench}: Holistic and Contamination Free Evaluation of Large Language Models for Code},
  author = {Jain, Naman and Han, King and Gu, Alex and Li, Wen-Ding and Yan, Fanjia and Zhang, Tianjun and Wang, Sida and Solar-Lezama, Armando and Sen, Koushik and Stoica, Ion},
  year = {2024},
  eprint = {2403.07974},
  archivePrefix = {arXiv},
  primaryClass = {cs.SE},
  url = {https://arxiv.org/abs/2403.07974}
}

@misc{guo2024deepseekcoder,
  title = {{DeepSeek-Coder}: When the Large Language Model Meets Programming---The Rise of Code Intelligence},
  author = {Guo, Daya and Zhu, Qihao and Yang, Dejian and Xie, Zhenda and Dong, Kai and Zhang, Wentao and Chen, Guanting and Bi, Xiao and Wu, Y. and Li, Y. K. and Luo, Fuli and Xiong, Yingfei and Liang, Wenfeng},
  year = {2024},
  eprint = {2401.14196},
  archivePrefix = {arXiv},
  primaryClass = {cs.SE},
  url = {https://arxiv.org/abs/2401.14196}
}

@inproceedings{rao2023iterative,
  title = {What Makes it Ok to Set a Fire? Iterative Self-distillation of Contexts and Rationales for Disambiguating Defeasible Social and Moral Situations},
  author = {Rao, Kavel and Jiang, Liwei and Pyatkin, Valentina and Gu, Yuling and Tandon, Niket and Dziri, Nouha and Brahman, Faeze and Choi, Yejin},
  booktitle = {Findings of the Association for Computational Linguistics: EMNLP 2023},
  year = {2023},
  doi = {10.18653/v1/2023.findings-emnlp.812},
  url = {https://aclanthology.org/2023.findings-emnlp.812/}
}

@inproceedings{zhang2025scoder,
  title = {{SCoder}: Progressive Self-Distillation for Bootstrapping Small-Scale Data Synthesizers to Empower Code {LLMs}},
  author = {Zhang, Xinyu and Zhou, Changzhi and Hu, Linmei and Zhang, Luhao and Chen, Xiancai and Fu, Haomin and Yang, Yang and Zhang, Mengdi},
  booktitle = {Findings of the Association for Computational Linguistics: EMNLP 2025},
  year = {2025},
  pages = {20825--20841},
  doi = {10.18653/v1/2025.findings-emnlp.1136},
  url = {https://aclanthology.org/2025.findings-emnlp.1136/}
}

@misc{sang2026crisp,
  title = {{CRISP}: Compressed Reasoning via Iterative Self-Policy Distillation},
  author = {Sang, Hejian and Xu, Yuanda and Zhou, Zhengze and He, Ran and Wang, Zhipeng and Sun, Jiachen},
  year = {2026},
  eprint = {2603.05433},
  archivePrefix = {arXiv},
  url = {https://arxiv.org/abs/2603.05433v7}
}

@misc{nicolicioiu2026diversity,
  title = {On-Policy Self-Distillation with Sampled Demonstrations Reduces Output Diversity},
  author = {Nicolicioiu, Andrei Liviu and Pezeshki, Mohammad and Courville, Aaron},
  year = {2026},
  eprint = {2606.26091},
  archivePrefix = {arXiv},
  url = {https://arxiv.org/abs/2606.26091v1}
}

@misc{gemmateam2025gemma3,
  title = {{Gemma 3} Technical Report},
  author = {{Gemma Team} and Kamath, Aishwarya and Ferret, Johan and Pathak, Shreya and Vieillard, Nino and Merhej, Ramona and Perrin, Sarah and Matejovicova, Tatiana and others},
  year = {2025},
  eprint = {2503.19786},
  archivePrefix = {arXiv},
  primaryClass = {cs.CL},
  url = {https://arxiv.org/abs/2503.19786},
  doi = {10.48550/arXiv.2503.19786}
}

@misc{yang2025qwen3,
  title = {{Qwen3} Technical Report},
  author = {Yang, An and Li, Anfeng and Yang, Baosong and Zhang, Beichen and Hui, Binyuan and Zheng, Bo and Yu, Bowen and Gao, Chang and Huang, Chengen and Lv, Chenxu and Zheng, Chujie and Liu, Dayiheng and Zhou, Fan and Huang, Fei and Hu, Feng and Ge, Hao and Wei, Haoran and Lin, Huan and Tang, Jialong and Yang, Jian and Tu, Jianhong and Zhang, Jianwei and Yang, Jianxin and Yang, Jiaxi and Zhou, Jing and Zhou, Jingren and Lin, Junyang and Dang, Kai and Bao, Keqin and Yang, Kexin and Yu, Le and Deng, Lianghao and Li, Mei and Xue, Mingfeng and Li, Mingze and Zhang, Pei and Wang, Peng and Zhu, Qin and Men, Rui and Gao, Ruize and Liu, Shixuan and Luo, Shuang and Li, Tianhao and Tang, Tianyi and Yin, Wenbiao and Ren, Xingzhang and Wang, Xinyu and Zhang, Xinyu and Ren, Xuancheng and Fan, Yang and Su, Yang and Zhang, Yichang and Zhang, Yinger and Wan, Yu and Liu, Yuqiong and Wang, Zekun and Cui, Zeyu and Zhang, Zhenru and Zhou, Zhipeng and Qiu, Zihan},
  year = {2025},
  eprint = {2505.09388},
  archivePrefix = {arXiv},
  primaryClass = {cs.CL},
  url = {https://arxiv.org/abs/2505.09388},
  doi = {10.48550/arXiv.2505.09388}
}
\clearpage
\appendix

\section{Metrics, estimators, and probability analysis}
\label{app:metrics}
\subsection{Correctness and conditional implementation distributions}
For a fixed task, let $V=v_x(Y)$ and $J=\phi(Y)$. Equation~\eqref{eq:factor} is simply $\Pr(V=1,J=j)=\Pr(V=1)\Pr(J=j\mid V=1)$. If $a=0$, unconditional correct richness is zero and $q$ is undefined. An observed pool with no successes need not have $a=0$; its matched-correct estimator is unavailable. For countably many classes the occupancy sums converge, because $1-(1-p)^k\le kp$ gives $C_k\le ka$ and $D_b\le b$. These quantities count expected observed classes at the stated budget, not all possible implementations.

\paragraph{Proof of Proposition~\ref{prop:breadth}.}
Let $I_j$ indicate observing at least one correct output from class $j$ in $k$ independent draws. Its expectation is $1-(1-aq_j)^k$. Linearity of expectation gives $C_k$. The same argument for $b$ draws from the conditional correct distribution yields $D_b$. All $k$ draws fail with probability $(1-a)^k$, so pass@$k$ depends only on $a$. For $q=(1,0,\ldots)$, $D_b=1$. For uniform mass on $J\ge2$ classes, $D_b=J[1-(1-1/J)^b]>1$ when $b\ge2$. Hence identical correctness at every budget does not identify conditional breadth.\hfill$\square$

More generally, spreading mass from a larger to a smaller conditional class, without reversing their order, cannot decrease $D_b$: $f(q)=1-(1-q)^b$ is concave for $b\ge2$. The same applies to $C_k$ at fixed $a$ and $k\ge2$. This describes a change in a probability distribution; it does not assert that the spectral operator induces that change in every network.

For $a>0$, the two richness endpoints are linked by the number of successful draws. Let $B\sim\operatorname{Binomial}(k,a)$ and set $D_0(q)=0$. Conditioning on $B$ gives
\begin{equation}
C_k(a,q)=\mathbb E\bigl[D_B(q)\bigr]
=\sum_{b=0}^k\binom kb a^b(1-a)^{k-b}D_b(q).
\label{eq:budget-correct-count-link}
\end{equation}
Given $B=b$, the $b$ correct classes are drawn from $q$, so their expected richness is $D_b(q)$. Thus total-budget richness combines how many correct samples are obtained with how broadly those samples are distributed. Matching the correct-draw budget isolates the latter quantity on the eligible tasks.

\subsection{Finite-sample estimators}
Suppose a task has $n$ generated outputs, $m$ of which pass its verifier, and correct-class counts $m_j$ with $\sum_jm_j=m$. We compute
\begin{align}
\widehat{\mathrm{pass@}k}&=1-\frac{\binom{n-m}{k}}{\binom nk}, &&1\le k\le n,\label{eq:passest}\\
\widehat C_k&=\sum_j\left[1-\frac{\binom{n-m_j}{k}}{\binom nk}\right], &&1\le k\le n,\label{eq:cest}\\
\widehat D_b&=\sum_j\left[1-\frac{\binom{m-m_j}{b}}{\binom mb}\right], &&1\le b\le m.\label{eq:dest}
\end{align}
We set $\binom uv=0$ for nonnegative integers $u<v$. Equation~\eqref{eq:passest} is the standard pass@$k$ estimator \citep{chen2021humaneval}. Equations~\eqref{eq:cest}--\eqref{eq:dest} are rarefaction estimators \citep{hurlbert1971nonconcept}. A uniformly chosen subset of $k$ outputs misses correct class $j$ with probability $\binom{n-m_j}{k}/\binom nk$. Summing its complement gives the exact expected subset richness conditional on the recorded pool; under independent sampling it estimates the corresponding population quantity. Conditioning on correct outputs gives the analogous calculation for $D_b$.

At $k=n$, $\widehat C_n$ is the number of observed correct classes. At $k=1$, $\widehat C_1=m/n=\widehat{\mathrm{pass@}1}$. At $b=1$, $D_1=1$ wherever defined. Thus neither $C_1$ nor $D_1$ distinguishes how probability is distributed among multiple correct classes. The plotted budgets are those present in the archived estimates; we do not reconstruct a missing pass@32 value by interpolation.

\subsection{AST fingerprints and concentration}
Python programs are extracted using the recorded first-fenced-block rule. The primary fingerprint parses the extracted program and conservatively normalizes its AST. Formatting and docstrings are removed; local identifiers are normalized only when scope handling is unambiguous. Function names, external identifiers, literals, and uncertain scopes are retained. Consequently, distinct fingerprints can reflect structural implementation variants within one algorithm. Conversely, a fingerprint does not establish functional equivalence beyond the recorded tests. Parse failures do not silently become new correct classes. Secondary exact-program and control-flow fingerprints provide different granularities.

An additional recorded concentration statistic is correct-output Simpson diversity,
\begin{equation}
\widehat S=1-\frac{\sum_j m_j(m_j-1)}{m(m-1)},\qquad m\ge2.
\end{equation}
It estimates the probability that two independently drawn correct outputs fall in different classes. Its aggregate uses tasks with enough correct samples and complete labels. It complements richness by emphasizing concentration, while sharing the limitations of the class definition. Exact-program, AST, and control-flow metrics need not move equally; the coarser control-flow comparisons do not establish a uniform increase in algorithmic families.

\subsection{Task aggregation and uncertainty}
\begin{figure}[htbp]
\centering\includegraphics[width=\linewidth]{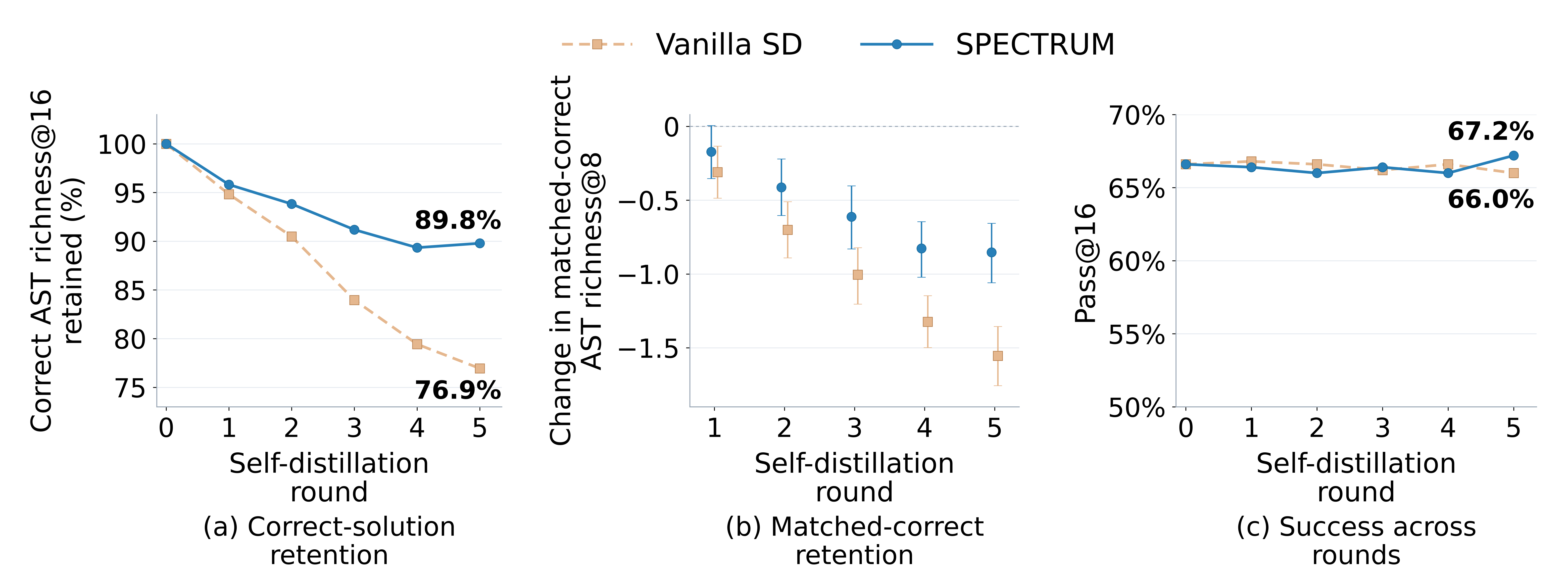}
\caption{\textbf{Within-correct retention over successive updates.} Original 16-sample MBPP comparison of Vanilla SD and \method. (a) $C_{16}$ relative to initialization; (b) paired $D_8$ changes from initialization on each contrast's common eligible tasks, with pointwise 95\% task-bootstrap intervals; (c) pass@16. Panels (a,c) show point estimates. The complete three-design comparison is Figure~\ref{fig:trajectories}.}
\label{fig:paired-trajectories}
\end{figure}

Pass@$k$ and $C_k$ are macro-averaged over the full task set. A marginal $D_b$ mean uses $\mathcal E_b(M)=\{x:m_x(M)\ge b\text{ and all correct outputs have AST labels}\}$ for model $M$ and reports its size. A paired contrast between $M_1,M_2$ averages task-level differences on $\mathcal E_b(M_1)\cap\mathcal E_b(M_2)$. Subtracting marginal means with different eligible sets is not this paired effect. Figure~\ref{fig:paired-trajectories}b pairs each round with initialization at $b=8$; its eligible population can change between rounds. It is a pointwise trajectory, not a single fixed cohort.

The five-round MBPP intervals use 2,000 task-bootstrap resamples. Transfer intervals also use 2,000 resamples, pairing by exact task ID and conditioning on common eligibility. Percentile endpoints are the 2.5th and 97.5th percentiles. The analysis script records the means, intervals, populations, and input hashes. These pointwise intervals describe evaluation-task uncertainty conditional on the saved training runs, not variation across independent training runs; no multiplicity adjustment is applied.

\FloatBarrier
\section{Operator derivation and mathematical scope}
\label{app:proofs}
\subsection{Reference loss and positionwise sensitivity}
Let $\mathcal T_i$ contain the nonpadding completion targets of reference $i$. The scalar differentiated during calibration is
\begin{equation}
\ell_i(\theta)=-\frac{1}{|\mathcal T_i|}\sum_{u\in\mathcal T_i}
\log p_\theta(y_{iu}^\star\mid x_i,y_{i,<u}^\star).
\label{eq:calibration-loss-appendix}
\end{equation}
This loss is computed separately for each reference before collecting gradients. For selected module $\ell$, let $h_{iu}^{(\ell)}\in\R^d$ be its native linear output at position $u$, written as a column here, and define
$g_{iu}^{(\ell)}=\nabla_{h_{iu}^{(\ell)}}\ell_i(\theta_t)$.
The second moment in Equation~\eqref{eq:geometry} averages the outer products of these gradients over every nonpadding position, including prompt and zero-gradient positions. Completion masking selects the loss targets; it does not prevent prompt activations from influencing subsequent completion predictions.

The statistic has a direct directional interpretation. For a unit vector $v$, perturbing one selected output position by $\epsilon v$ gives
\begin{equation}
\ell_i\bigl(h_{iu}^{(\ell)}+\epsilon v\bigr)
-\ell_i\bigl(h_{iu}^{(\ell)}\bigr)
=\epsilon\,g_{iu}^{(\ell)\top}v+o(\epsilon),
\end{equation}
where the notation denotes an intervention on that intermediate output, with the remaining computation propagated normally. Consequently,
\begin{equation}
v^\top C_{t,\ell}v
=\frac{1}{M_\ell}\sum_{i,u,\mathrm{nonpad}}
\bigl(g_{iu}^{(\ell)\top}v\bigr)^2.
\label{eq:directional-sensitivity-appendix}
\end{equation}
This averages squared first-order responses to individual positionwise perturbations. It does not include the cross-position terms arising from a coherent simultaneous perturbation of several outputs. Opposite gradient signs contribute positively, so $C_{t,\ell}$ measures sensitivity rather than a signed descent direction. It is an uncentered gradient second moment, not a Hessian of the reference loss.

When $\lambda_{\max}(C_{t,\ell})>0$, normalization produces
$0\preceq\Cbar_{t,\ell}\preceq\I$ and a maximum eigenvalue of one. Empty or zero-signal calibration is rejected because the normalization would otherwise be undefined. The same anchor is reused, but the gradients and normalized statistic are recomputed at the current student parameters.

\subsection{Proof of Proposition~\ref{prop:prox}: proximal solution and bounds}
For fixed $\Cbar$ and row activation $h$, write the objective in Equation~\eqref{eq:prox} as
\begin{equation}
F_h(z)=\frac12\|z-h\|_2^2+
\frac\tau2 z(\I-\Cbar)z^\top.
\end{equation}
Its Hessian is $H=\I+\tau(\I-\Cbar)$. If $\mu_j\in[0,1]$ are the eigenvalues of $\Cbar$, the eigenvalues of $H$ are
$1+\tau(1-\mu_j)\in[1,1+\tau]$.
Thus $F_h$ is strongly convex and has a unique minimizer. Its first-order condition is
\begin{equation}
z-h+\tau z(\I-\Cbar)=0,
\qquad zH=h.
\end{equation}
It follows that $z=hH^{-1}=hT_\tau$. Diagonalizing $H$ gives
\begin{equation}
T_\tau
=U\operatorname{diag}\!\left(
\frac{1}{1+\tau(1-\mu_j)}\right)U^\top,
\end{equation}
which is Equation~\eqref{eq:gain}.

All gains are positive at finite $\tau$, and lie in
$[1/(1+\tau),1]$. Because $T_\tau$ is symmetric positive definite, its singular values are precisely these gains. For any row vector $\delta$,
\begin{equation}
\frac{\|\delta\|_2}{1+\tau}
\le\|\delta T_\tau\|_2
\le\|\delta\|_2.
\end{equation}
Moreover,
\begin{equation}
\|T_\tau-\I\|_2
=\max_j\frac{\tau(1-\mu_j)}{1+\tau(1-\mu_j)}
\le\frac{\tau}{1+\tau}.
\end{equation}
These establish Equation~\eqref{eq:bound}. The condition number is at most $1+\tau$, and $T_0=\I$.

\subsection{Calibration stability}
For normalized statistics $A,B$ with spectra in $[0,1]$, write
$H_A=\I+\tau(\I-A)$ and $H_B=\I+\tau(\I-B)$.
The inverse identity gives
\begin{align}
T_\tau(A)-T_\tau(B)
&=H_A^{-1}(H_B-H_A)H_B^{-1}\\
&=\tau T_\tau(A)(A-B)T_\tau(B).
\end{align}
Both outer factors have operator norm at most one. Submultiplicativity therefore yields
\begin{equation}
\|T_\tau(A)-T_\tau(B)\|_2
\le\tau\|A-B\|_2.
\label{eq:normalized-stability-appendix}
\end{equation}
The argument requires neither an eigengap nor shared eigenvectors. This completes the proof of Proposition~\ref{prop:prox}.\hfill$\square$

The result concerns normalized statistics. To relate it to raw second moments, let $C,D\succeq0$ with
$c=\lambda_{\max}(C)\ge\gamma>0$ and
$d=\lambda_{\max}(D)\ge\gamma>0$.
Since $\|D\|_2=d$ and $|c-d|\le\|C-D\|_2$,
\begin{align}
\left\|\frac{C}{c}-\frac{D}{d}\right\|_2
&\le\frac{\|C-D\|_2}{c}
+\|D\|_2\left|\frac1c-\frac1d\right|\\
&=\frac{\|C-D\|_2+|d-c|}{c}
\le\frac{2\|C-D\|_2}{\gamma}.
\label{eq:raw-normalization-stability}
\end{align}
Combining the two estimates gives
\begin{equation}
\left\|T_\tau\!\left(\frac{C}{c}\right)
-T_\tau\!\left(\frac{D}{d}\right)\right\|_2
\le\frac{2\tau}{\gamma}\|C-D\|_2.
\end{equation}
This is a deterministic perturbation bound under a nonvanishing normalization scale. No sampling-concentration claim for the finite-anchor estimator is required.

\subsection{Relative reference sensitivity}
For a nonzero row vector $h$, define its sensitivity per unit squared norm as
\begin{equation}
s_{\Cbar}(h)=\frac{h\Cbar h^\top}{\|h\|_2^2}.
\end{equation}
Equation~\eqref{eq:relative-sensitivity} states that modulation cannot decrease this quantity. To prove it, write
$\Cbar=U\operatorname{diag}(\mu_j)U^\top$ and set
\begin{equation}
w_j=\frac{(hu_j)^2}{\|h\|_2^2},
\qquad
r_j=\frac{1}{[1+\tau(1-\mu_j)]^2}.
\end{equation}
The weights sum to one and $r_j>0$. Direct substitution yields
\begin{align}
s_{\Cbar}(hT_\tau)-s_{\Cbar}(h)
&=\frac{\mathbb E_w[\mu r]}{\mathbb E_w[r]}
-\mathbb E_w[\mu]\\
&=\frac{\operatorname{Cov}_w(\mu,r)}{\mathbb E_w[r]}\\
&=\frac{\sum_{i,j}w_iw_j(\mu_i-\mu_j)(r_i-r_j)}
{2\mathbb E_w[r]}
\ge0.
\label{eq:relative-sensitivity-proof}
\end{align}
The final inequality follows because $r_j$ is nondecreasing in $\mu_j$. For $\tau>0$, it is strictly increasing, so equality holds exactly when all eigendirections carrying nonzero energy from $h$ share the same eigenvalue. At $\tau=0$, modulation is the identity.

The relative increase does not require an increase in absolute sensitivity energy. Indeed, since $0<r_j\le1$,
\begin{equation}
(hT_\tau)\Cbar(hT_\tau)^\top
=\sum_j\mu_jr_j(hu_j)^2
\le\sum_j\mu_j(hu_j)^2
=h\Cbar h^\top.
\end{equation}
Thus modulation does not increase activation norm or absolute sensitivity energy, while sensitivity per unit squared norm does not decrease. This is a property of a local quadratic statistic; it does not assert a decrease in the original reference loss.

\subsection{Folding, finite precision, and the projection control}
For row activations and a linear module with weight $W$ and bias $b$,
\begin{equation}
(xW^\top+b^\top)T
=x(T^\top W)^\top+(T^\top b)^\top.
\end{equation}
Thus replacing $W$ and $b$ by $T^\top W$ and $T^\top b$ implements the same local map. This is the algebra underlying temporary weight folding. The implementation computes products in float32 and casts back to the model dtype; the exact invertibility statements concern real arithmetic. The insertion point is the native K/V linear output before rotary position operations and any subsequent key normalization. Grouped-query dimensions come from each selected module's actual output dimension. The original weights are saved and restored before learning.

The projection design uses $P_r=U_{1:r}U_{1:r}^\top$, with eigenvectors ordered by decreasing eigenvalue and $r$ equal to half the module output dimension in the completed run. It preserves retained directions and nulls the remainder. At $\tau=1$, \method's gains instead lie in $[1/2,1]$. The two designs differ in spectrum and effective strength, so this comparison does not match their distance from the identity.

For completeness, the infinite-strength limit of the proximal family is
\begin{equation}
\lim_{\tau\to\infty}T_\tau
=\sum_{j:\mu_j=1}u_ju_j^\top.
\end{equation}
It projects onto the maximum-eigenvalue eigenspace of the normalized statistic, not an arbitrary prescribed top-$r$ space. Finite-strength full rank is therefore the relevant property for the completed experiment. 

\subsection{Local operator properties and learned solution distributions}
Positive gains preserve distinctions between inputs to the local linear map. Subsequent attention, nonlinear layers, vocabulary projection, stochastic decoding, and finite-data learning determine whether those distinctions affect output classes. The operator results establish bounded modulation, calibration stability, and relative sensitivity enrichment. The measurements of $C_k$ and $D_b$ establish how the resulting students distribute their correct outputs.

A separate idealization clarifies why learning from self-generated sequences is not automatically an improvement step. For a fixed prompt $x$, consider the completion-only, unnormalized sequence-NLL objective
\begin{equation}
L_t(\theta)=-\mathbb E_{Y\sim p_{\theta_t}(\cdot\mid x)}
\bigl[\log p_\theta(Y\mid x)\bigr].
\end{equation}
When the sequence distribution is normalized and differentiation can be interchanged with summation, its gradient at the generating parameters is
\begin{equation}
\left.\nabla_\theta L_t(\theta)\right|_{\theta=\theta_t}
=-\sum_yp_{\theta_t}(y\mid x)
\left.\nabla_\theta\log p_\theta(y\mid x)\right|_{\theta=\theta_t}
=-\left.\nabla_\theta\sum_yp_\theta(y\mid x)\right|_{\theta=\theta_t}
=0.
\end{equation}
This identity assumes exact model-distribution sampling and the complete sequence log likelihood. The recorded learner in Equation~\eqref{eq:student} includes prompt targets and uses example-wise token averaging; a sequence-dependent length normalization need not preserve the zero-gradient identity. The practical loop also uses a decoding distribution, finite corpora, a fixed optimization schedule, and, for \method, reference-calibrated generation. Its improvement and retention trajectories are measured empirically.

\FloatBarrier
\section{Algorithm, information flow, and protocol}
\label{app:protocol}
\subsection{A complete round}
The following sequence implements a complete round of \method, separating the temporary generator from the native learner.
\begin{enumerate}
\item Fix synthesis prompts $\cQ$, reference anchor $\cA$, held-out evaluation tasks, decoding settings, and the initial checkpoint $\theta_0$ before the loop.
\item At round $t$, calculate reference-completion loss under $\theta_t$, collect selected K/V-output gradients, and form $\Cbar_{t,\ell}$ on the unchanged anchor.
\item Construct $T_{t,\ell}$ with the fixed strength $\tau$ and temporarily fold the operators into the current checkpoint's native linear weights.
\item Sample $S$ completions for each synthesis prompt. Keep every raw record. No execution outcome, reward, uniqueness judgment, or AST label determines membership or weight in $\cD_t$.
\item Restore the original $\theta_t$ weights. Fit one LoRA adapter on $\cD_t$, merge it into the base weights, and save $\theta_{t+1}$.
\item Evaluate $\theta_{t+1}$ through common native inference, storing all outputs and verification labels. These labels do not affect the next training round. Repeat from step 2 using $\theta_{t+1}$.
\end{enumerate}
For Vanilla SD or SSD, the same generate--learn--update loop uses the corresponding standard or decoding-shaped policy, omitting reference calibration and weight folding.
The anchor itself is never expanded with newly judged samples. It is reused as supervision at every calibration step, so ``fixed'' describes the information source rather than an initialization-only operation. In the completed study, there is no paired intervention-on generator diagnostic at every round; the native endpoints establish retained differences after learning, not the fraction of an immediate generator benefit transferred by SFT.

\subsection{Data and prompts}
The MBPP preparation starts from the full official splits. Exact normalized prompt overlaps with held-out prompts are removed before partitioning the training pool; the recorded excluded training IDs are 602, 704, and 872. The resulting sizes are 291 synthesis, 50 calibration, 30 validation, and 500 evaluation tasks. A Python signature stub is extracted mechanically from the reference AST to specify the interface. Its implementation and evaluation assertions are not placed in the synthesis/evaluation prompt. The calibration partition supplies complete reference solutions.

The fixed MBPP prompt body is:
\begin{quote}\small\ttfamily
[task description]\\[3pt]
Implement this Python function. Return only the complete implementation.\\
{[signature stub]}
\end{quote}
Brackets identify fields in a template, not missing manuscript results. The same template is used across methods; it does not request multiple strategies or select a strategy. Model-native chat formatting is applied by the tokenizer. Prompt and token provenance are stored in the experiment records.

HumanEval+ follows the recorded EvalPlus task interface and full base-plus tests, using release v0.1.10 with EvalPlus 0.3.1. APPS Intro is a fixed 200-task evaluation subset using the stored function-interface adapter. For stdin/stdout tasks, the prompt requests \texttt{solve(stdin: str) -> str}; the evaluator checks the returned output string against supplied cases using whitespace-token equality. This adapted interface, task subset, and evaluation budget are part of the result and differ from official benchmark leaderboard settings. No HumanEval+ or APPS references recalibrate the frozen MBPP students.

\begin{table}[t]
\centering\small
\caption{\textbf{Completed training and evaluation settings.} The final 64-sample MBPP supplement is a separate sample pool; only 16-sample measurements exist for all historical rounds.}
\label{tab:settings}
\begin{tabular}{p{.42\linewidth}p{.51\linewidth}}
\toprule
Component & Recorded setting\\
\midrule
Backbone & Qwen/Qwen2.5-Coder-1.5B-Instruct; BF16, SDPA\\
Rounds / records per round & 5 / $291\times16=4,656$\\
Calibration & 50 fixed references; completion-only mean NLL; length limit 1,536\\
Geometry & Middle and last decoder layers; K/V outputs before RoPE; full second moment\\
\method\ / projection control & $\tau=1$ / rank fraction $0.5$\\
LoRA & Rank 8, alpha 8, dropout 0.05; attention Q/K/V/O projections\\
SFT & 1 epoch; learning rate $10^{-5}$; weight decay 0.01; warm-up ratio 0.03; gradient clipping 1.0\\
SFT batching / length & Microbatch 1, accumulation 16; maximum 1,536 tokens\\
SFT target mask & All nonpadding causal targets, including prompt targets\\
Common generation & Temperature 0.8, top-$p$ 0.95, top-$k$ 0\\
SSD training generation & Temperature 1.5, top-$p$ 0.8, top-$k$ 20\\
MBPP generation limits & 512 new tokens, 1,024 prompt tokens\\
Transfer generation limits & 1,024 new tokens, 4,096 prompt tokens\\
Native evaluation draws & MBPP every round: 16; final supplement: 64; transfer: 16\\
Uncertainty & 2,000 task-bootstrap resamples; one training run\\
\bottomrule
\end{tabular}
\end{table}

\subsection{What the student optimizes}
Let $s$ denote the concatenated tokenized prompt and generated completion, and let $\mathcal T(s)$ contain its nonpadding causal targets. The recorded microbatch-one schedule accumulates example-level token-mean losses,
\begin{equation}
\ell_s(\theta)=-\frac{1}{|\mathcal T(s)|}\sum_{u\in\mathcal T(s)}\log p_\theta(s_u\mid s_{<u}),
\qquad \mathcal L_{\mathrm{SFT}}=\frac{1}{|\cD_t|}\sum_{s\in\cD_t}\ell_s(\theta).
\end{equation}
Gradient accumulation follows the recorded minibatch schedule. Including prompt targets and averaging each example by its own length are explicit choices, distinct from completion-only calibration and globally token-weighted corpus optimization. Every record receives the same outer example weight; token normalization within each record follows its own length. Correctness and source subspace do not determine training weights. Truncation at the fixed context limit is a sequence-length constraint, not quality filtering. The completed training summaries record zero truncated training examples.

\subsection{Information budgets and comparators}
Vanilla SD and SSD do not use the reference anchor. \method\ and the projection control use the same 50 references at each calibration step. Hence their direct comparison controls access to calibration information, whereas comparing either to Vanilla SD or SSD includes that difference in information use. SSD here adapts the published decoding idea to the small-model MBPP budget; it does not reproduce the original larger-scale results. UA-RL, which uses rewards for rare strategies, is discussed as related work but is not a completed numerical comparator in this archive.

\FloatBarrier
\section{Complete measured results}
\label{app:results}
\subsection{Longitudinal MBPP results at 16 samples}
Table~\ref{tab:rounds} contains all recorded rounds. The initial SSD row is an independently sampled evaluation of the same starting model, retained separately from the original trajectory. Small differences in initialization estimates are sampling/protocol-stream differences, not different learned checkpoints. No 64-sample intermediate-round results are inferred from these rows.

\begin{table}[t]
\centering\small
\caption{\textbf{Every completed native-student round on MBPP, 16 draws per task.} Full 500-task means for pass@1 and $C_{16}$; marginal $D_4$ means with eligibility. SSD is the separate continuation.}
\label{tab:rounds}
\begin{tabular}{llrrrr}
\toprule
Method & Round & pass@1 (\%) & $C_{16}$ & $D_4$ & Eligible\\
\midrule
Initial & 0 & 37.92 & 4.016 & 3.310 & 262\\
\midrule
Vanilla SD & 1 & 38.42 & 3.808 & 3.173 & 261\\
Vanilla SD & 2 & 39.15 & 3.634 & 3.080 & 269\\
Vanilla SD & 3 & 39.65 & 3.372 & 2.895 & 257\\
Vanilla SD & 4 & 40.79 & 3.190 & 2.803 & 270\\
Vanilla SD & 5 & 41.30 & 3.090 & 2.747 & 268\\
\midrule
\method & 1 & 37.84 & 3.848 & 3.222 & 261\\
\method & 2 & 38.36 & 3.768 & 3.159 & 259\\
\method & 3 & 39.05 & 3.662 & 3.063 & 262\\
\method & 4 & 39.78 & 3.588 & 2.991 & 263\\
\method & 5 & 40.18 & 3.606 & 2.997 & 263\\
\midrule
Projection control & 1 & 37.92 & 3.692 & 3.144 & 257\\
Projection control & 2 & 39.20 & 3.592 & 3.059 & 267\\
Projection control & 3 & 39.40 & 3.314 & 2.889 & 256\\
Projection control & 4 & 41.09 & 3.162 & 2.774 & 266\\
Projection control & 5 & 41.35 & 3.066 & 2.690 & 262\\
\midrule
SSD initial & 0 & 37.81 & 3.992 & 3.311 & 260\\
SSD & 1 & 37.80 & 3.846 & 3.232 & 260\\
SSD & 2 & 36.81 & 3.710 & 3.200 & 250\\
SSD & 3 & 36.60 & 3.724 & 3.196 & 256\\
SSD & 4 & 37.09 & 3.752 & 3.181 & 257\\
SSD & 5 & 35.96 & 3.710 & 3.268 & 255\\\bottomrule\end{tabular}\end{table}

\subsection{Paired retention trajectories}

\subsection{Final projection comparison and correct-count matching}
\begin{table}[htbp]
\centering\small
\caption{\textbf{Final 64-sample projection ablation.} Same 500 tasks, fixed reference access, and five-round training schedule. Marginal $D_4$ means are descriptive; eligible tasks are 321 and 323.}
\begin{tabular}{lrrrrr}
\toprule
Generator & pass@1 (\%) & pass@64 (\%) & $C_{64}$ & Retention (\%) & $D_4$\\
\midrule
Subspace projection &41.77&70.20&8.388&65.5&2.810\\
\method &40.33&72.60&11.510&89.9&3.143\\
\bottomrule
\end{tabular}
\end{table}

\begin{table}[htbp]
\centering\small
\caption{\textbf{Paired final MBPP richness contrasts, \method\ minus Vanilla SD.} Common eligible tasks are selected separately for each correct-draw budget. Intervals are the archived 95\% task-bootstrap intervals.}
\begin{tabular}{lrrr}
\toprule
Endpoint & Paired difference & 95\% interval & Tasks\\
\midrule
$C_{64}$ &3.004&$[2.568,3.494]$&500\\
$D_4$ &0.295&$[0.249,0.343]$&316\\
$D_8$ &0.780&$[0.665,0.889]$&296\\
$D_{16}$ &1.816&$[1.582,2.055]$&254\\
\bottomrule
\end{tabular}
\end{table}
Sparse $D_{64}$ estimates are available on few fully correct tasks but do not form a primary claim: the paired \method--Vanilla SD population is only 29 tasks. At the recorded 16-sample round-five endpoint, the \method-minus-projection difference in exact-program $D_4$ is $0.312$ with interval $[0.253,0.377]$, and the AST difference is $0.293$ $[0.233,0.353]$. The coarser control-flow difference is $0.042$ $[-0.011,0.095]$. All three contrasts use 251 common eligible tasks and distinguish implementation structure from coarser control-flow variation.

\subsection{Transfer contrasts and the projection control}
\begin{table}[htbp]
\centering\small
\caption{\textbf{Frozen-student transfer after MBPP learning.} Sixteen samples per task, without transfer training or recalibration. Success and $C_{16}$ average over all tasks. $D_4$ is the marginal mean over each row's eligible tasks; paired common-task contrasts appear in the text. Higher values indicate greater success or richness; eligibility is descriptive.}
\label{tab:transfer-full}
\begin{tabular}{@{}llrrrrr@{}}
\toprule
Benchmark & Student & pass@1 (\%) & pass@16 (\%) & $C_{16}$ & $D_4$ & Eligible\\
\midrule
\multirow{3}{*}{HumanEval+ (164)} & Initial &53.09&85.37&5.768&3.410&116\\
 & Vanilla SD &53.47&86.59&5.848&3.422&120\\
 & \method &51.94&86.59&6.000&3.497&114\\
\midrule
\multirow{3}{*}{APPS Intro (200)} & Initial &15.47&41.50&2.070&3.772&50\\
 & Vanilla SD &15.19&35.50&1.930&3.632&46\\
 & \method &13.97&34.00&1.845&3.721&43\\
\bottomrule
\end{tabular}
\end{table}

\begin{table}[htbp]
\centering\small
\caption{\textbf{Paired transfer effects of \method\ relative to Vanilla SD.} Differences use all tasks for pass@1 and $C_{16}$, and common eligible tasks for $D_4$. Intervals use 2,000 paired task-bootstrap resamples of the saved task records. Students receive no transfer training or recalibration.}
\begin{tabular}{llrrr}
\toprule
Benchmark & Endpoint & Difference & 95\% interval & Tasks\\
\midrule
HumanEval+ & pass@1 (pp)&$-1.524$&$[-3.088,0.076]$&164\\
 & $C_{16}$&0.152&$[-0.128,0.439]$&164\\
 & $D_4$&0.080&$[0.021,0.142]$&111\\
APPS Intro & pass@1 (pp)&$-1.219$&$[-1.937,-0.530]$&200\\
 & $C_{16}$&$-0.085$&$[-0.210,0.055]$&200\\
 & $D_4$&0.104&$[0.007,0.205]$&42\\
\bottomrule
\end{tabular}
\end{table}
The projection control's HumanEval+ scores are pass@1 $52.74\%$, pass@16 $84.76\%$, $C_{16}=5.921$, and $D_4=3.473$ on 115 eligible tasks. Its APPS Intro scores are $14.88\%$, $34.50\%$, $1.880$, and $3.685$ on 47 eligible tasks. The \method-minus-projection paired $D_4$ effects are $0.039$ $[-0.021,0.102]$ on 109 HumanEval+ tasks and $0.067$ $[-0.003,0.142]$ on 42 APPS tasks. The paired intervals specify the precision of these smaller transfer contrasts.

\FloatBarrier
\section{Cross-model and cross-benchmark baseline context}
\label{app:matrix}
The completed matrix evaluates Qwen2.5-Coder-1.5B-Instruct,
DeepSeek-Coder-6.7B-Instruct \citep{guo2024deepseekcoder}, Gemma-3-4B-IT \citep{gemmateam2025gemma3}, and Qwen3-8B \citep{yang2025qwen3} on
MBPP, HumanEval+, APPS Intro, CodeContests \citep{li2022alphacode}, and LiveCodeBench \citep{jain2024livecodebench}.
Table~\ref{tab:matrix} gives pass@16; Table~\ref{tab:matrix-ci} reports
the archived task-bootstrap intervals. These are initial-model
evaluations, independent of the sample pools used for the five-round
and frozen-student comparisons. Consequently, even the same initial
checkpoint can have different finite-sample estimates across tables.

All twenty cells completed generation and verification. Task counts
are 500, 164, 200, 165, and 200, respectively. MBPP pass@16 is
estimated from 64 samples per task; the other benchmarks use 16.
HumanEval+ uses official EvalPlus verification; APPS Intro,
CodeContests, and LiveCodeBench use the recorded native adapted
verifiers. The Gemma--LiveCodeBench
run uses eager attention. These protocol choices define the reported
benchmark setting.

\begin{table}[htbp]
\centering\small
\setlength{\tabcolsep}{5pt}
\caption{\textbf{Completed cross-model baseline matrix.} Pass@16 (\%)
for initial models across five coding benchmarks. Each column has a
common recorded evaluation setting across models. These rows measure
initial-model performance, not trained \method\ students.}
\label{tab:matrix}
\begin{tabular}{@{}lrrrrr@{}}
\toprule
Model & MBPP & HumanEval+ & APPS Intro & CodeContests & LiveCodeBench\\
\midrule
Qwen2.5-Coder-1.5B & 66.51 & 85.98 & 36.50 & 0.61 & 22.00\\
DeepSeek-Coder-6.7B & 67.48 & 87.80 & 50.50 & 6.06 & 22.50\\
Gemma-3-4B & 58.07 & 73.78 & 38.00 & 3.03 & 24.00\\
Qwen3-8B & 62.38 & 93.29 & 51.00 & 5.45 & 40.50\\
\bottomrule
\end{tabular}
\end{table}

\begin{table}[htbp]
\centering\small
\caption{\textbf{Archived uncertainty for the baseline matrix.}
Pointwise 95\% task-bootstrap intervals for pass@16 on the percentage
scale, conditional on the recorded model and sample pool.}
\label{tab:matrix-ci}
\begin{tabular}{@{}llrr@{}}
\toprule
Model & Benchmark & pass@16 (\%) & 95\% interval\\
\midrule
Qwen2.5-Coder-1.5B & MBPP & 66.51 & $[62.83,70.31]$\\
 & HumanEval+ & 85.98 & $[80.49,90.85]$\\
 & APPS Intro & 36.50 & $[30.00,43.00]$\\
 & CodeContests & 0.61 & $[0.00,2.42]$\\
 & LiveCodeBench & 22.00 & $[16.50,28.00]$\\
\midrule
DeepSeek-Coder-6.7B & MBPP & 67.48 & $[63.69,71.39]$\\
 & HumanEval+ & 87.80 & $[82.91,92.68]$\\
 & APPS Intro & 50.50 & $[44.00,57.00]$\\
 & CodeContests & 6.06 & $[2.42,9.70]$\\
 & LiveCodeBench & 22.50 & $[17.00,28.50]$\\
\midrule
Gemma-3-4B & MBPP & 58.07 & $[53.64,62.21]$\\
 & HumanEval+ & 73.78 & $[67.06,79.88]$\\
 & APPS Intro & 38.00 & $[31.50,44.50]$\\
 & CodeContests & 3.03 & $[0.61,6.06]$\\
 & LiveCodeBench & 24.00 & $[18.50,30.00]$\\
\midrule
Qwen3-8B & MBPP & 62.38 & $[58.31,66.59]$\\
 & HumanEval+ & 93.29 & $[89.02,96.95]$\\
 & APPS Intro & 51.00 & $[44.00,58.00]$\\
 & CodeContests & 5.45 & $[2.42,9.09]$\\
 & LiveCodeBench & 40.50 & $[33.50,47.50]$\\
\bottomrule
\end{tabular}
\end{table}

\FloatBarrier
\section{Actual equal-correct-count examples}
\label{app:cases}
\begin{figure}[htbp]
\centering\figfile{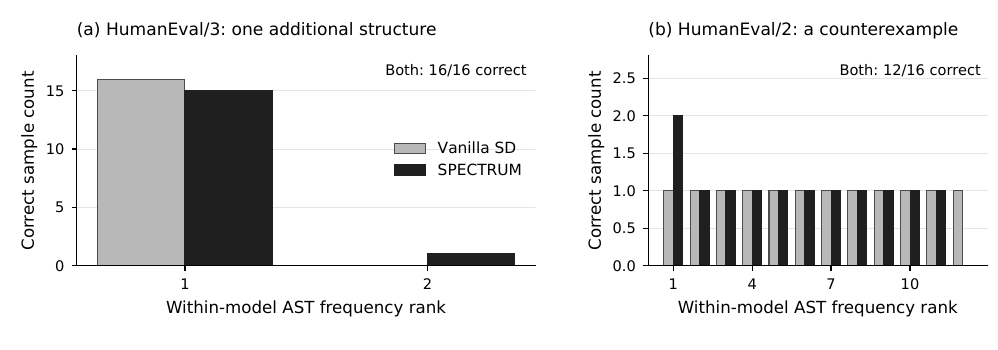}
\caption{\textbf{The same number of correct outputs can conceal different structures.} Actual saved round-five HumanEval+ samples. (a) Both methods pass all 16 draws on HumanEval/3. Vanilla SD has one AST class with count 16; \method\ has counts 15 and 1, giving $\widehat D_4=1$ and $1.25$. (b) Both pass 12 of 16 draws on HumanEval/2; Vanilla SD has twelve singleton classes, while \method\ has one doubleton and ten singletons. Their $\widehat D_4$ values are $4$ and $3.909$. Classes are ranked separately within each method, so horizontal position does not assert matching class identity. Both cases follow the stated lowest-ID rule.}
\label{fig:examples}
\end{figure}

The compact archive retains correctness and program fingerprints but removes program text. We therefore show actual class-frequency records rather than reconstructing purported code samples. For each sign of the \method-minus-Vanilla SD observed AST-count difference, choose the numerically lowest HumanEval task ID with equal correct counts of at least four. This rule yields HumanEval/3 for a positive example and HumanEval/2 for a negative example (Figure~\ref{fig:examples}). Selection conditions on the sign and is illustrative, not an estimate of prevalence.

These examples instantiate the measurement distinction without supplying semantic algorithm labels. At equal observed $n$ and $m$, every finite-sample pass@$k$ estimator agrees between methods, while the correct-class occupancy can differ in either direction. Equal estimated pass curves are not proof that the underlying true success probabilities are identical.

\FloatBarrier
\section{Resources and realized sample lengths}
\label{app:resources}
All methods use the same 23,280 generated training records over five rounds. Table~\ref{tab:resources} records their realized lengths. \method\ generates $9.9\%$ more completion tokens than Vanilla SD; equal record count is therefore not equal token count or FLOPs. Matching the number of correct evaluation samples does not control training length or compute. It answers a distributional question about the saved students.

\begin{table}[htbp]
\centering\small
\caption{\textbf{Training-generation resources across five rounds.} Counts are summed from saved round budgets; mean length is generated tokens divided by records. Length-capped records remain in training. SFT token counts include prompt targets under the recorded loss mask.}
\label{tab:resources}
\begin{tabular}{lrrrrr}
\toprule
Method & Records & Generated tokens & Mean length & Capped & SFT tokens\\
\midrule
Vanilla SD &23,280&4,859,928&208.76&617&6,373,288\\
SSD &23,280&5,812,400&249.67&1,498&7,325,760\\
\method &23,280&5,342,368&229.48&833&6,855,728\\
Projection control &23,280&4,727,089&203.05&619&6,240,449\\
\bottomrule
\end{tabular}
\end{table}

For a selected module of output dimension $d$ and $M$ gradient positions, dense second-moment accumulation costs $O(Md^2)$ and eigendecomposition costs $O(d^3)$, excluding the reference forward/backward passes. Dense folding of a $d\times p$ weight costs $O(d^2p)$. These operations are performed during calibration/construction, not at each generation token. The final student has the original dense architecture. Archived runtime records cover different hardware and stages; we make no matched wall-clock or GPU-memory efficiency claim from their combination.

\FloatBarrier
\section{One-round self-distillation study}
\label{app:pilot}
This experiment uses Qwen2.5-Coder-1.5B-Instruct with 128 synthesis tasks, one generated completion per task, one generate--learn cycle, and five SFT epochs. Evaluation uses 64 held-out MBPP tasks and 64 samples per task. Five SFT epochs within a cycle are not five self-distillation rounds. Its task counts and training schedule differ from the main five-round study, so the two experiments are reported separately.

\paragraph{Code extraction and verification.}
For each saved completion, we extract the first Python, \texttt{py}, or untagged fenced code block and execute it against the task tests. Surrounding explanatory text is ignored. We re-evaluate all four generation designs with this same extraction rule.

\begin{table}[htbp]
\centering\small
\caption{\textbf{Complete one-round comparison.} One generate--learn cycle, 64 MBPP tasks, 64 samples each. Values are archived rounded summaries; bold marks the largest estimate at shown precision. AST and exact-program statistics are structural proxies.}
\label{tab:pilot-full}
\begin{tabular}{@{}lrrrr@{}}
\toprule
Metric $\uparrow$ & Vanilla SD & SSD & Projection & \method\\
\midrule
Correct samples / 4,096 &1,508&\textbf{1,511}&1,510&1,501\\
pass@1 (\%) &36.82&\textbf{36.89}&36.87&36.65\\
pass@8 (\%) &56.49&\textbf{56.79}&55.78&56.18\\
pass@32 (\%) &61.61&61.66&60.84&\textbf{63.11}\\
pass@64 (\%) &62.50&62.50&60.94&\textbf{65.63}\\
Solved tasks / 64 &40&40&39&\textbf{42}\\
Exact-program unique fraction &0.802&0.797&0.796&\textbf{0.814}\\
AST unique fraction &0.622&0.622&0.616&\textbf{0.649}\\
AST Simpson diversity &0.864&0.863&0.869&\textbf{0.872}\\
Correct-count-matched AST $D_4$ &3.393&3.391&3.409&\textbf{3.422}\\
\bottomrule
\end{tabular}
\end{table}

\begin{table}[htbp]
\centering\small
\caption{\textbf{Paired one-round effects relative to SSD.} \method\ minus SSD, with pointwise task-bootstrap intervals for the evaluated checkpoints.}
\label{tab:pilot-paired}
\begin{tabular}{@{}lrr@{}}
\toprule
Endpoint & Difference & 95\% interval\\
\midrule
Correct-count-matched AST $D_4$ &$+0.0312$&$[0.0095,0.0550]$\\
Exact-program unique fraction &$+0.0083$&$[0.0009,0.0171]$\\
AST Simpson diversity &$+0.0094$&$[0.0034,0.0162]$\\
pass@1 (pp) &$-0.244$&$[-0.854,0.342]$\\
pass@64 (pp) &$+3.125$&$[0,7.813]$\\
\bottomrule
\end{tabular}
\end{table}

\begin{figure}[htbp]
\centering\includegraphics[width=\linewidth]{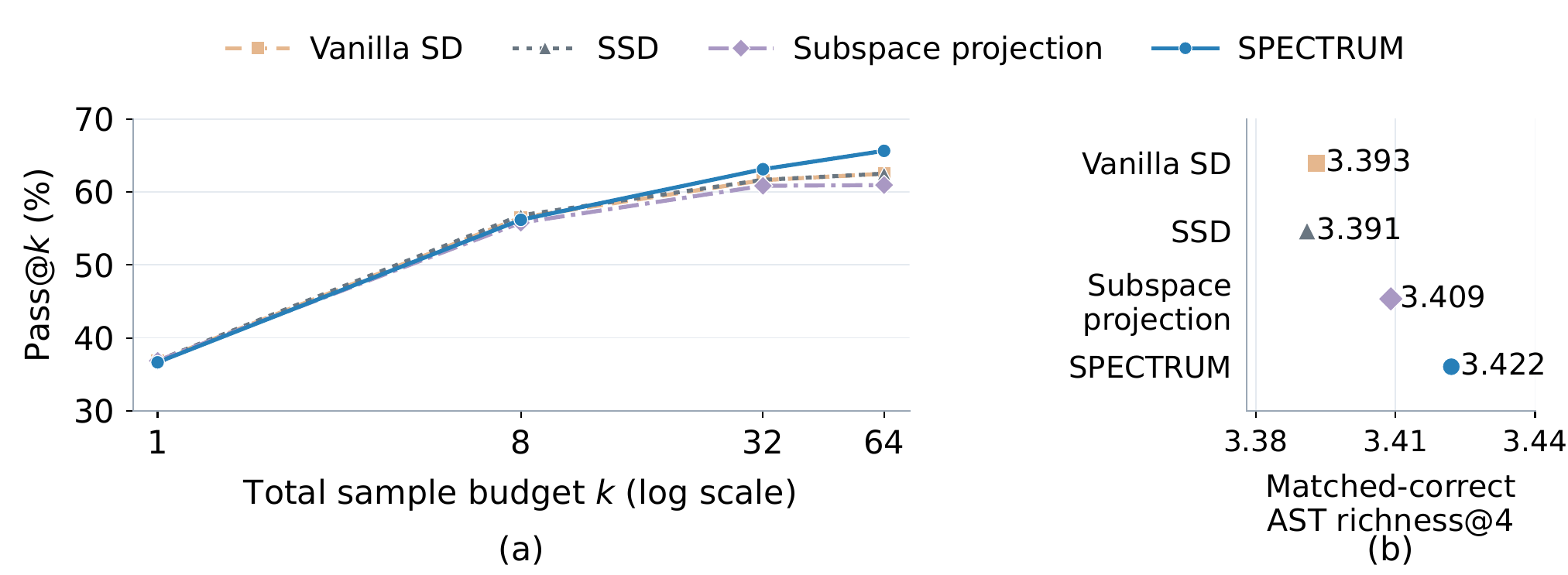}
\caption{\textbf{Complementary evidence from the one-round study.} (a) Pass@$k$ at recorded sampling budgets; (b) correct-count-matched AST $D_4$. All four generation designs are included. Marks use archived rounded point estimates from 64 tasks and 64 samples per task; marginal error bars are unavailable. Table~\ref{tab:pilot-paired} gives the recorded paired uncertainty. This protocol is separate from the five-round study.}
\label{fig:pilot}
\end{figure}

The one-round study combines higher high-budget success and matched-correct richness than SSD with slightly lower single-draw correctness. The full five-round experiment subsequently measures a different operating point: \method\ has higher correctness than SSD, while SSD has higher $C_{16}$ and marginal $D_4$. Different synthesis budgets, task sets, and optimization schedules preclude attributing this difference solely to loop length.

\FloatBarrier
\section{Relation to adjacent self-distillation settings}
\label{app:positioning}
\paragraph{Sequence-level and token-feedback distillation.}
Our learner uses sampled sequences as cross-entropy targets. Demonstration-conditioned on-policy distillation instead supplies dense teacher feedback conditioned on privileged context. \method\ shapes generation through fixed reference calibration, then trains one native student on its raw corpus. Looped Self-Distillation names the outer generate--learn--update process.

\paragraph{Iterative self-improvement.}
Iterative context/rationale learning, code-data synthesis, and teacher refresh have been studied with different feedback mechanisms \citep{rao2023iterative,zhang2025scoder,sang2026crisp}. Our study combines a fixed-information protocol, a continuous spectral generation intervention, and longitudinal measurements of conditional correct-implementation richness. Demonstration-conditioned diversity analyses \citep{nicolicioiu2026diversity} address a different teacher-feedback construction; the present results measure contraction and retention through repeated learning from unselected raw corpora.

\end{document}